%% file: main.tex
\documentclass[10pt,twocolumn,letterpaper]{article}

\PassOptionsToPackage{table}{xcolor}
\usepackage[pagenumbers]{cvpr}

\input{preamble}

\definecolor{cvprblue}{rgb}{0.21,0.49,0.74}
\usepackage[pagebackref,breaklinks,colorlinks,allcolors=cvprblue]{hyperref}

\title{TokenDial: Continuous Attribute Control for Text-to-Video \\ Generation in Visual Dial Space}

\author{
\begin{tabular}{cccc}
Zhixuan Liu$^{1,2,*}$ & Peter Schaldenbrand$^{2}$ & Yijun Li$^{1}$ & Long Mai$^{1}$ \\
Aniruddha Mahapatra$^{1}$ & Cusuh Ham$^{1}$ & Jean Oh$^{2}$ & Jui-Hsien Wang$^{1}$
\end{tabular}\\[0.8em]
$^1$Adobe Research \qquad
$^2$Carnegie Mellon University \\[0.6em]
\url{https://tokendial.github.io}
}
\date{}

\begin{document}

\twocolumn[{%
\renewcommand\twocolumn[1][]{#1}
\maketitle
\vspace{-0.8cm}
\begin{center}
  \vspace{-0.5cm}
  \centering\small
  \captionsetup{type=figure}
  \includegraphics[width=\textwidth]{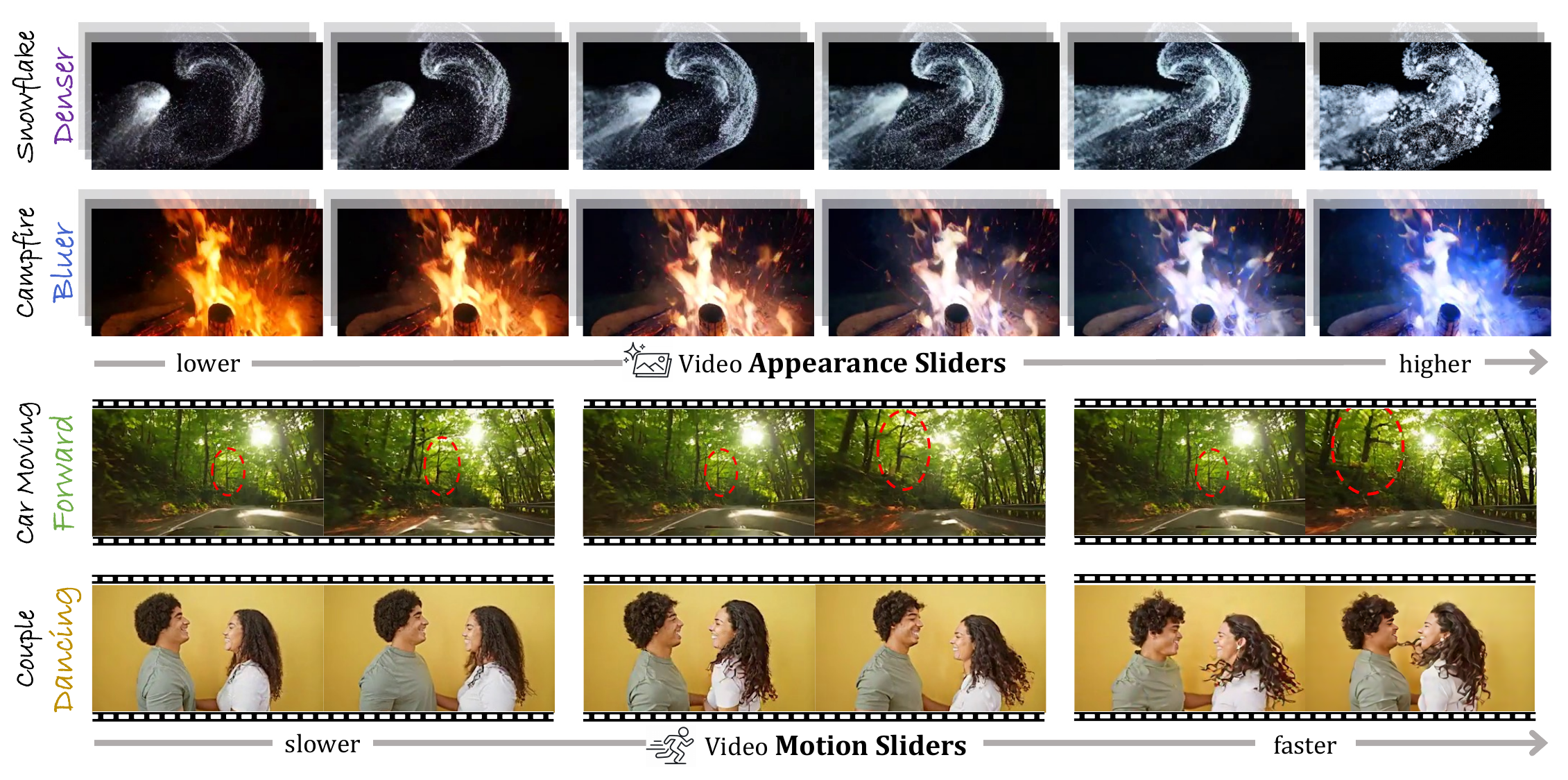}%
  \caption{\textbf{\emph{TokenDial} enables continuous visual dials for controlling both appearance and motion in text-to-video generation.}
Increasing the dial strength produces smooth, monotonic changes in appearance attributes (top) and motion dynamics (bottom), while preserving the underlying content and temporal coherence.
For motion sliders, red circles highlight the same spatial region at matched time steps to visualize the change in motion magnitude.
Left-side attribute labels are shown for illustration only and are not used as prompts.}
  \label{fig:teaser}
\end{center}
}]

\begingroup
\renewcommand\thefootnote{*}
\footnotetext{Work was done during an internship at Adobe Research}
\endgroup

\begin{abstract}

In video diffusion transformers, visual patch tokens maintain explicit correspondence to space and time. We hypothesize that their channel dimension can serve as a semantic control space, which we call Visual Dial Space $V^+$. In this space, additive directions can be broadcast to the token stream to control appearance or motion attributes, enabling slider-style edits such as making a generated person look older or run faster. To verify the hypothesis, we present \projname, a framework for learning attribute directions in the proposed $V^+$ space. 
\projname~keeps the pretrained video generator frozen and optimizes only additive directions. Rather than requiring paired edited videos, \projname~supervises each direction through its induced effect on generated videos: the edited video should move along the desired attribute while the remaining content stays stable. The learned directions become reusable visual dials that support continuous appearance and motion control, explicit spatiotemporal localization, composition, and reuse across prompts, resolutions, and video lengths. Experiments and human studies show that \projname~achieves stronger slider controllability and better content preservation than prior video editing and slider-based methods.

\end{abstract}

\input{sections/1_intro}
\input{sections/2_related_work}
\input{sections/3_Vplus_space}

\input{sections/4_discover_attribute}
\input{sections/5_inference_vplus_space}

\input{sections/5_experiment}

\input{sections/6_new_results}

\input{sections/7_conclusion}

\section*{Acknowledgments}
We thank Xuanlei Zhao and Rui Chen for their helpful discussion. This work was supported in part by NSF IIS-2112633.

{
  \small
  \bibliographystyle{ieeenat_fullname}
  \bibliography{main_bibliography}
}

\clearpage
\appendix
\startcontents[appendix]
\begingroup
\renewcommand{\tableofcontents}{%
  \section*{Contents}
  \printcontents[appendix]{}{1}{}
}
\input{supplimentary/sections/x_suppl_new}
\endgroup
\stopcontents[appendix]

\end{document}

%% file: preamble.tex
\usepackage{url}
\usepackage{pifont}
\usepackage{makecell}
\usepackage{float}
\usepackage{titletoc}
\usepackage{listings}
\usepackage{placeins}

\newcommand{\projname}{\textit{TokenDial}}
\newcommand{\vnamecapital}{Visual Dial Space}

\newcommand{\greencheck}{\textcolor{green}{\ding{51}}}
\newcommand{\redcross}{\textcolor{red}{\ding{55}}}
\newcommand{\gc}{\greencheck}
\newcommand{\rc}{\redcross}

\providecommand{\Description}[1]{}

\lstdefinestyle{vlmprompt}{
  basicstyle=\ttfamily\scriptsize,
  breaklines=true,
  columns=fullflexible,
  keepspaces=true,
  frame=single,
  framerule=0.3pt,
  xleftmargin=0.5em,
  framexleftmargin=0.5em,
  aboveskip=0.6em,
  belowskip=0.6em,
  showstringspaces=false
}

%% file: sections/1_intro.tex
\section{Introduction}
\label{sec:intro}

Text-to-video (T2V) generation has advanced rapidly, producing high-quality videos from high-level prompts. Yet for real creative workflows, realism is only the starting point: creators need the ability to keep a scene fixed while precisely controlling how much a particular attribute changes. Prompts can specify what appears (e.g., ``a campfire'', ``a person''), but they are a weak interface for dialing attribute strength (e.g., fire intensity, aging, or motion magnitude) without drifting identity, background, or temporal coherence. We study continuous semantic scalar control: slider-style modulation of continuous attributes while preserving the rest of the video.

Existing methods only partially address this need. Video-to-video (V2V) editing can modify video content through progressive prompts, but does not provide continuous, monotonic control over attribute strength. Recent video sliders and image-slider methods adapted to video improve fine-grained control, yet they primarily target appearance attributes and leave motion dynamics, such as intensity, rhythm, or perceived speed, difficult to dial. Moreover, many slider methods encode edits through text conditioning or lightweight weight updates (e.g., LoRA \cite{hu2022lora}), which makes the spatiotemporal extent of the edit implicit: whether an attribute changes in a particular region or moment is largely decided by the model rather than specified by the user. 
Table~\ref{tab:design_requirements} summarizes this gap: practical video dials should control \emph{what} changes, how much it changes, and \emph{where} and \emph{when} the change applies, while preserving identity, background, and temporal coherence.

This motivates us to look inside the visual representation of video
generators. A video diffusion transformer (DiT)~\cite{peebles2023scalable} operates on a visual
patch-token space $V$, where a video is represented as a sequence of
tokens with explicit correspondence to space and time. This token
structure is an appealing substrate for control: unlike text prompts
or model weights, visual tokens are directly tied to where and when
content appears in the generated video. However, defining a control over the full token grid would tie it to a specific resolution and length, and would not provide a reusable semantic notion of an attribute. We instead ask whether
attribute control can be factorized: a compact direction describes
\emph{what} visual property to change, while the token grid specifies
\emph{where} and \emph{when} the change is applied.

We therefore introduce visual dial space $V^+$, a compact control
space over the channel dimension of the visual patch tokens. A direction
in $V^+$ can be broadcast to the visual token stream, scaled to vary
edit strength, and gated by a soft spatiotemporal mask. This yields a
simple control interface: the $V^+$ direction specifies \emph{what} attribute to change, the scalar strength specifies \emph{how much} to change, and the mask over $V$ specifies \emph{where} and \emph{when} the change applies.

\begin{figure}
  \centering
  \includegraphics[width=\columnwidth]{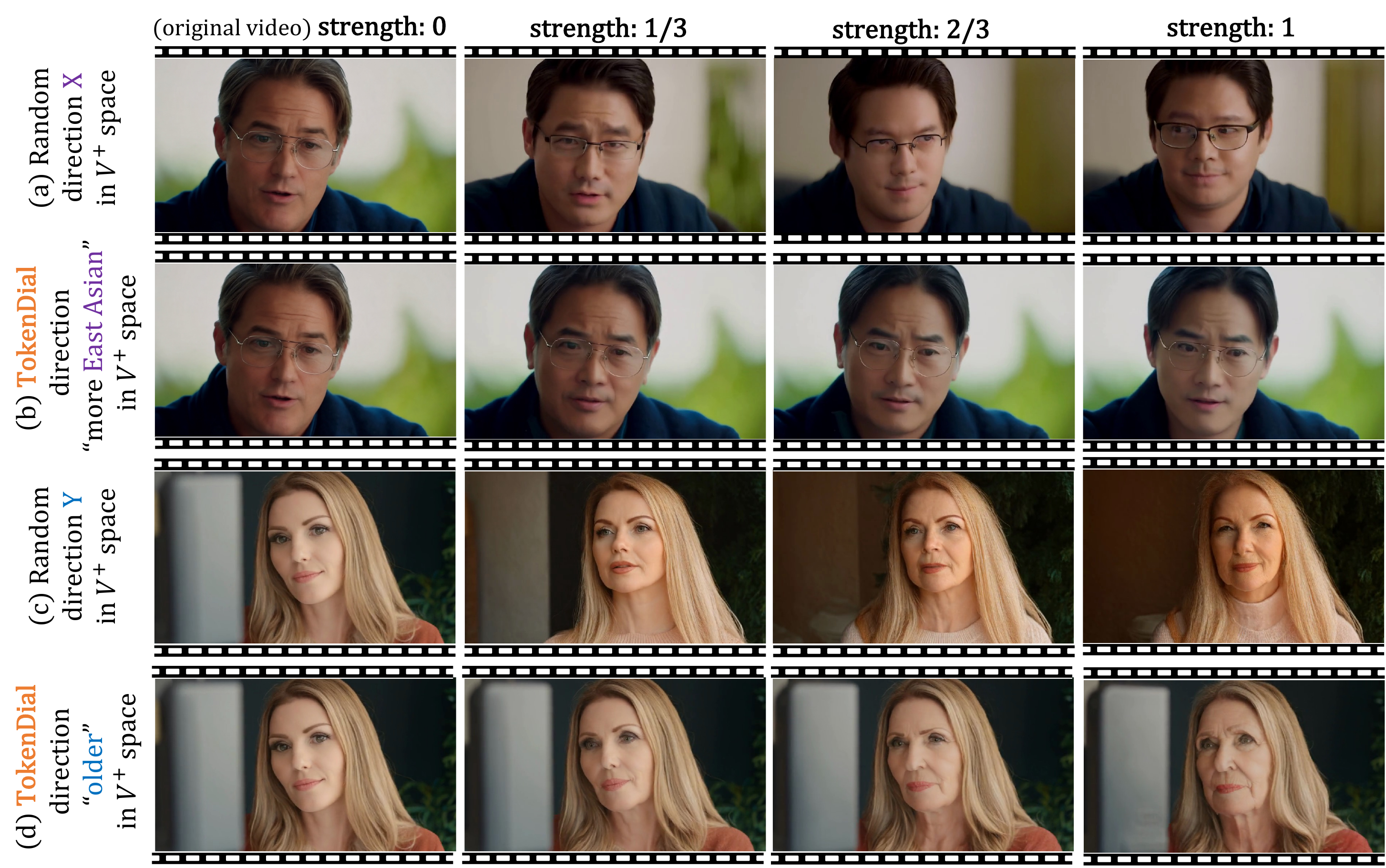}
  \caption{\textbf{Probing visual dial space.}
Some random directions in $V^+$ behave like semantic sliders when scaled, e.g., directions we interpret as ``more East Asian'' and ``older''. 
However, they are entangled with unintended changes to identity, eyeglasses, background, and motion. 
TokenDial learns cleaner attribute directions, yielding more targeted and better-preserved sliders.}
  \label{fig:motivation}
\end{figure}

Our motivation comes from a simple probing observation. 
We sample random directions in $V^+$, inject them into a frozen video DiT, and vary their scalar strength during generation.
Although $V^+$ presents the potential for semantically rich controllability, it is challenging to learn its semantic representation. 
As shown in Fig.~\ref{fig:motivation}, increasing the strength of one random direction gradually changes the facial appearance of a generated subject while roughly preserving the scene layout. This suggests that $V^+$ is not merely an arbitrary offset space: some directions already correspond to meaningful semantic changes. However, random directions are not usable controls. They are sparse, entangled, and found only by
chance; the facial-appearance change is coupled with unintended
changes to clothing, background, and motion, and users cannot specify the target attribute.
The key question is whether these semantic
directions can be learned systematically and aligned with user intent.

We present \projname, a framework for learning attribute directions
in $V^+$ space. \projname~keeps the pretrained video generator frozen and optimizes only compact additive directions injected into intermediate visual tokens.
Rather than requiring paired edited videos or updating
the backbone, \projname~learns a direction by supervising its induced
effect on generated videos: the edited video should move along the
desired attribute while the remaining content stays stable. We
instantiate this principle for both appearance and motion sliders.
For appearance, we align the generated change with semantic
directions in a pretrained video-understanding feature space. For
motion, we optimize feature-space motion objectives that scale
motion magnitude while preserving the motion pattern.

Once learned, a $V^+$ direction becomes a plug-and-play visual dial.
It can be scaled for continuous strength control, localized with spatiotemporal masks, reused across prompts, resolutions, and lengths supported by the same backbone, and composed with
other directions. This gives \projname~a unified interface for
continuous video control: a learned direction controls the attribute,
a scalar controls its strength, and a mask controls its spatial and
temporal support.

We validate \projname~on diverse appearance and motion attributes,
showing smooth and monotonic edits with strong preservation of
identity, background, and temporal coherence. Across metrics and human studies, \projname~outperforms prior video
sliders, image sliders adapted to video, text-driven video editing,
and image-to-video editing pipelines. 
Our contributions are threefold:
\begin{itemize}
    \item We introduce visual dial space, $V^+$, a compact control
    space over visual patch-token channels for continuous attribute
    control in pretrained video DiTs.
    \item We present \projname, a framework for
    learning $V^+$ directions for both appearance and motion
    control, without paired edited videos or backbone fine-tuning.
    \item We show that learned $V^+$ directions provide scalable,
    localizable, composable, and reusable dials,
    achieving stronger control, and identity and background preservation than prior slider
    and video editing methods.
\end{itemize}

%% file: sections/2_related_work.tex
\section{Related Work}
\label{sec:relwork}

\noindent\textbf{Controllable video generation.}
~Recent advances have enabled diverse forms of control in generative models, ranging from architecture- and conditioning-based designs~\cite{zhang2023controlnet,chen2024pixart,gal2023an} to application-driven controls such as subject personalization~\cite{ruiz2023dreambooth,hu2025hunyuancustom,garibi2025tokenverse} and camera motion guidance~\cite{xing2025motioncanvas,lee2026generative,geng2025motion,bai2025recammaster,mou2024revideo}. While these directions greatly improve usability, they typically address discrete or coarse controls (what to generate, where to place it, or which trajectory to follow), rather than continuously dialing the strength of an attribute during generation. In contrast, we study continuous, slider-style control of appearance and motion attributes in T2V models.

\noindent\textbf{Video editing.}
~Instruction- or example-based video editing methods~\cite{ju2026editverse,zi2025senorita,ku2024anyv2v,decart2025lucyedit,liu2025generative,wei2026univideo} achieve strong open-domain edits. However, their control interface is typically discrete: edit strength is specified through instruction text or a small set of pre-defined levels, which makes it difficult to achieve continuous, monotonic slider-style modulation of attribute strength. They also often rely on substantial edited supervision or synthetic training data, limiting scalability. In contrast, we target continuous attribute control during generation, using pretrained signals instead of paired edits.

\noindent\textbf{Fine-grained and continuous video generation.}
This line of work explores slider-based controls on attributes with various setups and architectures~\cite{gandikota2024concept,gandikota2025sliderspace,chiu2026text,ezra2025freesliders}. FreeSliders~\cite{ezra2025freesliders} and Text Slider~\cite{chiu2026text} have begun to study slider-style control for videos. FreeSliders provides training-free concept steering via an inference-time approximation of noise updates, whereas Text Slider learns lightweight LoRA directions in the shared text encoder for plug-and-play, reusable control. Other works in the image domain also pursue fine-grained, continuous edits by injecting LoRA modules or steering text/conditioning tokens~\cite{gandikota2024concept,gandikota2025sliderspace,kamenetsky2025saedit,parihar2025kontinuouskontextcontinuousstrength}. Recent work further improves continuous edit-strength
control by modifying classifier-free guidance~\cite{wolf2026continuouscontroleditingmodels}. Despite
controllability, existing methods largely keep the edit mechanism and its
spatiotemporal extent implicit, which limits precise localization, weakens identity
preservation, and makes motion dynamics difficult to dial.

\noindent\textbf{Semantics of various latent spaces.}
~Semantic structure in representation spaces has long enabled controllable generation, from linear directions in GAN latent space~\cite{radford2015unsupervised,chen2016infogan,harkonen2020ganspace} to semantic bottlenecks in diffusion models that support editing and interpretability~\cite{kwon2023diffusion,dalva2024noiseclr,park2023unsuperviseddiscoverysemanticlatent}. Recent work further shows that aligning denoising representations with pretrained understanding models improves training efficiency and generation quality, including extensions to video~\cite{yu2025repa,zhang2026videorepa,seo2026propflylearningpropagateonthefly}. 
TokenVerse~\cite{garibi2025tokenverse} uses the modulation space of Flux-like models for multi-concept personalization. 
Different from this model-specific space, TokenDial learns attribute directions in video DiT visual-token space, enabling explicit spatiotemporal control across different video backbones.

\input{tables/intro_table} 

%% file: tables/intro_table.tex
\begin{table}[t]
\centering
\scriptsize   
\caption{Capabilities explicitly supported or evaluated by baselines}
\label{tab:design_requirements}
\begin{tabular}{@{}c|cccc@{}}       
\toprule
       & Appear.\&        & Prog.   & Explicit & Identity  \\
Method & Motion           & Scaling & Mask     & Preserv.  \\
\midrule
I2I + V2V       & \rc & \rc & \gc & \gc \\
Text-based V2V  & \rc & \rc & \gc & \gc \\
Concept Sliders & \rc & \gc & \rc & \rc \\
SliderSpace     & \rc & \gc & \rc & \rc \\
Text Slider     & \rc & \gc & \rc & \rc \\
FreeSliders     & \rc & \gc & \rc & \rc \\
\textbf{\projname~(Ours)} & \gc & \gc & \gc & \gc \\
\bottomrule
\end{tabular}
\end{table}

%% file: sections/3_Vplus_space.tex
\begin{figure*}[tbp]
  \includegraphics[width=\textwidth]{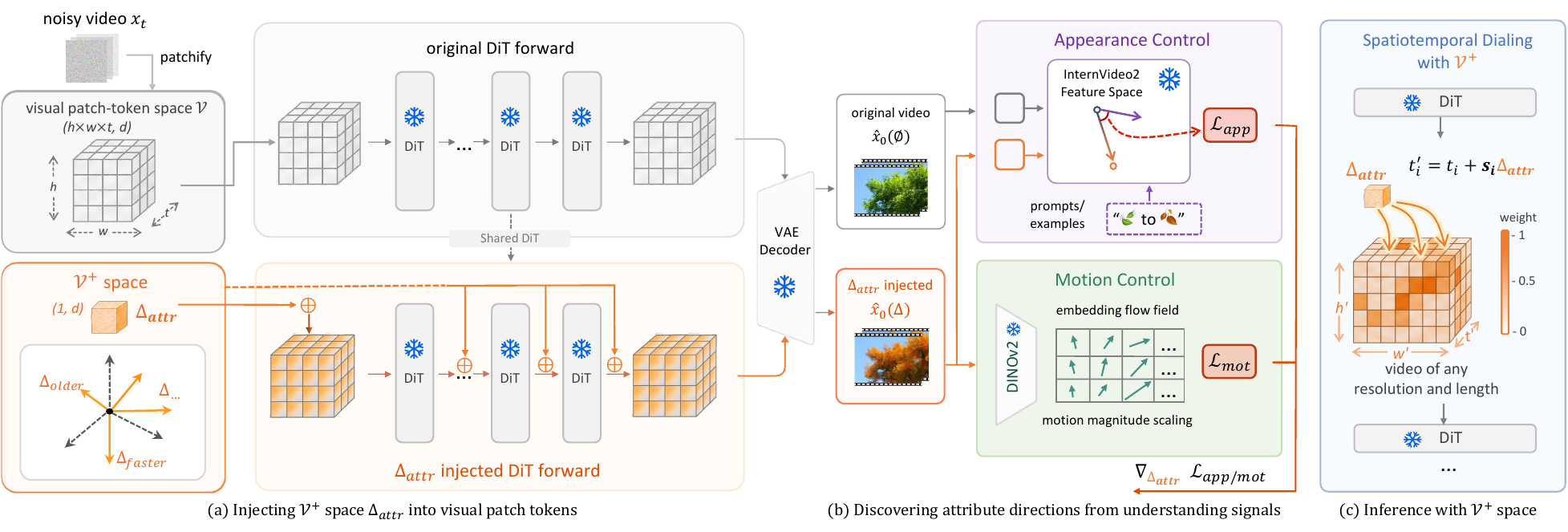}
    \caption{\textbf{Overview of \projname.} 
    \projname~learns visual dials in $V^+$, a compact channel-space offset over visual patch tokens of a frozen text-to-video DiT. 
    A direction $\Delta_{\text{attr}}$ (layer-specific) is broadcast to intermediate tokens and optimized through its induced video-level effect, using semantic direction matching for appearance and motion-magnitude scaling for motion. 
    At inference, the same $V^+$ direction can be scaled and gated by spatiotemporal masks for continuous, localized control across resolutions and video lengths.}
  \label{fig:method_overview}
  \Description{Overview diagram of the proposed method, showing learnable spatiotemporal token offsets in a video model.}
\end{figure*}

\section{\vnamecapital~for Video Control}
\label{sec:section_for_vplus}

\subsection{Preliminaries: Visual Patch-Token Space \texorpdfstring{$V$}{V}}
\label{sec:visual_token_space}

A pretrained DiT operates on a sequence of visual patch tokens. 
Given a latent video $x\in\mathbb{R}^{C\times F\times H\times W}$ encoded by a VAE encoder, the model applies a patchification operator $\mathrm{Patchify}(\cdot)$ to obtain
\begin{equation}
\mathcal{T}=\mathrm{Patchify}(x)\in\mathbb{R}^{L\times d},
\end{equation}
where $d$ is the hidden dimension and $L$ is the number of visual patch tokens, which depends on the video resolution and length. 
We denote the corresponding visual patch-token space as $V:=\mathbb{R}^{L\times d}$, and a token sequence as $\mathcal{T}\in V$.

Although $\mathcal{T}$ is written as a sequence, its tokens retain correspondence to the latent video grid: each token index $i\in\{1,\dots,L\}$ maps to a specific spatiotemporal region. 
Equivalently, the sequence can be viewed as a tensor in $\mathbb{R}^{f\times h\times w\times d}$, where $f,h,w$ are the token-grid dimensions over time, height, and width. 
This makes $V$ a natural interface for localizing edits. 
However, $V$ itself is not a compact attribute space: since $L$ changes with video resolution and length, a control defined directly in $\mathbb{R}^{L\times d}$ would be tied to a particular spatiotemporal grid.

\subsection{\vnamecapital~\texorpdfstring{$V^+$}{V+}}
\label{sec:v_plus_space}

We introduce a compact control space over the channel dimension of visual patch tokens:
\begin{equation}
    V^+ := \mathbb{R}^{d}.
\end{equation}
An element $\Delta \in V^+$ is an additive direction that can be broadcast to the visual patch-token sequence. 
Given visual patch tokens $\mathcal{T}=\{t_i\}_{i=1}^{L}$, we inject $\Delta$ as
\begin{equation}
    \mathcal{T}'=\{\, t_i + s_i\Delta \mid i=1,\dots,L \,\},
    \label{eq:vplus_injection}
\end{equation}
where $s=\{s_i\}_{i=1}^{L}$, $s_i\in[0,1]$, is a soft token mask over the visual patch tokens. 
The same channel direction is therefore shared across tokens, while the mask determines its spatiotemporal support on the token grid, and its scale controls the editing strength.

In our implementation, we use a layer-wise version of $\Delta$, since different transformer layers capture different levels of visual and temporal abstraction. 
Let $\Delta^{(k)} \in \mathbb{R}^{d}$ denote the offset injected after the $k$-th transformer block:
\begin{equation}
    \mathcal{T}'^{(k)} = \{\, t_i^{(k)} + s_i \Delta^{(k)} \mid i=1,\dots,L \,\}.
    \label{eq:layer_vplus_injection}
\end{equation}
We refer to the collection $\{\Delta^{(k)}\}$ as one $V^+$ direction for an attribute, and write it simply as $\Delta$ when the layer index is not essential. 

\subsection{Control Properties of \texorpdfstring{$V^+$}{V+}}
\label{sec:vplus_properties}
This construction gives $V^+$ three useful properties for video control. 
\textbf{(1) Grid-independent reusability.} 
A direction lies in $\mathbb{R}^{d}$ rather than $\mathbb{R}^{L\times d}$, so its parameterization is independent of the number of visual tokens and can be reused across prompts, resolutions, and video lengths supported by the same backbone. 
\textbf{(2) Spatiotemporal addressability.} 
After being broadcast to $V$, the same direction can be gated by token-level masks, enabling spatial and temporal localization without changing the direction. 
\textbf{(3) Continuous and compositional controllability.} 
Because directions are additive and share the same $V^+$ space, they can be smoothly scaled and linearly combined, e.g., $\lambda\Delta_a+\gamma\Delta_b$, to form compositional attribute controls.

These properties make $V^+$ a compact and addressable parameterization for video control, but they do not make every vector in $V^+$ meaningful. 
Arbitrary offsets may induce weak, entangled, or unintended changes. 
We therefore formulate video control as a direction-learning problem: given a target attribute, learn an offset direction $\Delta_{\mathrm{attr}}\in V^+$ whose video-level effect matches the desired change while preserving the rest of the scene.

%% file: sections/4_discover_attribute.tex
\section{Learning Attribute Directions in \texorpdfstring{$V^+$}{V+}}
\label{sec:learning_directions}
We now describe how TokenDial turns a target attribute into a visual dial.
Given an attribute such as ``larger'', ``spring to autumn'', or ``faster motion'', we keep the pretrained video generator frozen and optimize only additive direction
$
\Delta_{\text{attr}} \in {V}^{+} \;.
$
The direction is learned through its induced effect on generated videos: injecting $\Delta_{\text{attr}}$ should move the output along the desired attribute while leaving unrelated content stable.
We use pretrained understanding models to provide this video-level supervision, and instantiate the same direction-learning procedure for appearance and motion control.

\noindent\textbf{Posterior-refined supervision.}
Directly applying external supervision to a one-step estimate from an intermediate noise level can be unstable, since $\hat{x}_0$ reconstructed from a highly noisy $x_t$ often lacks reliable visual details. 
At each training iteration, we sample a timestep $t$, obtain $x_t$ from the clean latent $x_0$, and run the frozen denoiser twice: once without offsets and once with $\Delta_{\text{attr}}$ injected into the token stream.
This produces $\hat{x}_0(\varnothing)$ and $\hat{x}_0(\Delta)$ via Tweedie's formula~\cite{ho2020denoising}.
Before computing the understanding loss, we unroll $K$ additional reverse steps to obtain refined estimates $\hat{x}_0^{\text{ref}}(\varnothing)$ and $\hat{x}_0^{\text{ref}}(\Delta)$.
For efficiency, gradients are stopped through the refinement unroll and backpropagated only through the initial prediction, similar to~\cite{liu2024scoft}.
In our experiments, we use a small $K$ (e.g., $K{=}4$) for high-noise timesteps.

\noindent\textbf{Appearance dial: semantic direction matching.}
For appearance attributes, we supervise the edit in a pretrained video-understanding embedding space using InternVideo2~\cite{wang2024internvideo2}.
Let $\mathcal{E}(\cdot)$ denote the InternVideo2 visual encoder.
Given refined reconstructions with and without the offset, we define the predicted attribute direction as
\begin{equation}
    \mathbf{d}_{\text{pred}} = \mathcal{E}\!\left(\hat{x}_0^{\text{ref}}(\Delta)\right) - \mathcal{E}\!\left(\hat{x}_0^{\text{ref}}(\varnothing)\right).
\end{equation}
We align this direction with a target semantic direction $\mathbf{d}_{\text{tgt}}$ using cosine distance:
\begin{equation}
    \mathcal{L}_{\text{app}} = 1 - \cos\!\left(\mathbf{d}_{\text{pred}}, \mathbf{d}_{\text{tgt}}\right).
\end{equation}
To discourage unrelated changes, we add a perceptual preservation term between the edited and unedited reconstructions:
\begin{equation}
    \mathcal{L}_{\text{appear}} = \mathcal{L}_{\text{app}} + \lambda_a \cdot \mathrm{LPIPS}\!\left(\hat{x}_0^{\text{ref}}(\Delta), \hat{x}_0^{\text{ref}}(\varnothing)\right),
\end{equation}
where $\lambda_a$ controls the regularization strength.
We obtain $\mathbf{d}_{\text{tgt}}$ by contrasting either paired text prompts (e.g., ``hot'' vs.\ ``cold'') or exemplar videos, and projecting them into the same InternVideo2 feature space.

\noindent\textbf{Motion dial: motion magnitude scaling.}
For motion attributes, we supervise the magnitude of motion rather than static appearance.
We extract frame-wise patch embeddings using DINOv2~\cite{oquab2024dinov2} and compute optical flow on these embeddings with the Lucas--Kanade (LK) method~\cite{lucas1981}, yielding a motion field $\mathbf{m}(\cdot)$.
Our goal is to scale motion strength by a factor $\gamma$ ($\gamma>1$ amplifies motion; $\gamma<1$ attenuates it). 
A fixed reference motion field is not ideal, because the edited video may become temporally misaligned as its motion becomes faster or slower.
We therefore define a self-scaled target using the stop-gradient $.\mathrm{sg}()$ motion field of the edited sample:
\begin{equation}
    \mathcal{L}_{mot} = \| \mathbf{m}(\hat{x}_0^{ref}(\Delta)) - \gamma \cdot [\mathbf{m}(\hat{x}_0^{ref}(\Delta))\text{.sg( )}] \|_2^2.
\end{equation}
This objective changes the motion magnitude while avoiding frame-wise alignment to a separate reference trajectory.
To preserve appearance and identity cues, we additionally regularize DINOv2 features on the first frame:
\begin{equation}
\mathcal{L}_{\text{ff}}
=
1-\cos\!\left(
\mathcal{D}\!\left(\hat{x}_{0,\,t{=}0}^{\text{ref}}(\Delta)\right),
\mathcal{D}\left(\hat{x}_{0,t=0}^{ref}(\varnothing)\text{.sg( )}\right)
\right),
\end{equation}
where $\mathcal{D}(\cdot)$ denotes the DINOv2 feature extractor on a single frame.
The final motion objective is 
\begin{equation}
\mathcal{L}_{\text{motion}}=\mathcal{L}_{\text{mot}}+\lambda_m\,\mathcal{L}_{\text{ff}},
\end{equation}
where $\lambda_m$ controls the regularization strength.
This learns a $V^+$ direction that modulates motion magnitude while preserving the underlying motion pattern and identity cues.

%% file: sections/5_inference_vplus_space.tex
\section{Using \texorpdfstring{$V^+$}{V+} Directions for Video Control}
\label{sec:inference}

After training, a learned $V^+$ direction becomes an inference-time
visual dial. TokenDial exposes two complementary handles: a soft
token mask $\mathbf{s}$ that specifies the spatiotemporal locality support of
the edit, and an edit scale $s_{\text{edit}}$ that controls the global
slider strength.

\noindent\textbf{Spatiotemporal masking.}
We apply a learned direction $\Delta$ with a token-level soft mask
$\mathbf{s}\in[0,1]^L$, where $L$ is the number of visual patch tokens.
For each token, $s_i$ controls the local strength of the edit, also the
mask specifies both where and when the attribute change is applied.
The mask can be provided by the user, defined over spatial or temporal
regions, or automatically derived from attention maps of the target
text token. Since visual patch tokens preserve the video layout, these
masks act as soft spatiotemporal mattes over the video,
enabling localized edits, spatially and temporally varying
effects while reducing drift in unrelated regions.

\noindent\textbf{Global strength control.}
A direct way to obtain a slider is to scale the injected offset
$\Delta$. This works for moderate strengths, but large offset scales
can over-perturb the feature trajectory and degrade video quality.
We therefore control the slider strength at the vector-field level by
scaling only the flow component induced by the learned direction.
Let $\Phi_\theta$ denote the video model parameterizing the velocity field, and $\Phi_\theta(x_t,t,c)$ its predicted velocity under condition $c$, with $\varnothing$ denoting the unconditional input.
We define the update $\tilde{u}$ as
\begin{equation}
\begin{aligned}
\tilde{u}
=\, & \Phi_\theta(x_t,\varnothing)
+ s_{\text{txt}}\!\left(
    \Phi_\theta(x_t,c)
    - \Phi_\theta(x_t,\varnothing)
\right) \\
& + s_{\text{edit}}\!\left(
    \Phi_\theta(x_t,c,\Delta)
    - \Phi_\theta(x_t,c)
\right).
\end{aligned}
\end{equation}
The first two terms form the standard text-guided base trajectory,
while the last term isolates the edit-induced flow. Varying
$s_{\text{edit}}$ changes only the attribute-specific
component, producing continuous slider control while preserving the
text-consistent structure of the generated video. This keeps the slider path stable across strengths.

%% file: sections/5_experiment.tex
\section{Experimental Setup}
\label{sec:experiment}

\noindent\textbf{Tasks.}
We evaluate slider-style control over concept--attribute pairs: a \emph{concept} is the target entity or phenomenon, and an \emph{attribute} is a continuous property to be dialed, such as color, size, age, dilution, or motion magnitude. 
A slider is tested by generating videos from the same prompt and seed at increasing strengths, and evaluating whether the attribute changes progressively while identity, background, and temporal coherence are preserved. 
We study both appearance sliders and motion sliders, where motion sliders change motion magnitude without changing the underlying motion pattern.

\noindent\textbf{Dataset and protocol.}
We evaluate 12 concepts with 5 attributes each, covering object-centric appearance, volumetric effects, fluids, particle-like phenomena, and articulated motion. 
Appearance dials use small concept-specific text--video collections without paired edits, while motion dials are trained from a few hundred green-screen clips and applied universally at inference. 
For each concept--attribute pair, we generate 16 base videos and 5 slider strengths under matched prompts, seeds, and inference budgets, yielding roughly 4.8K videos per method. 
We also evaluate dials learned from synthetic backbone-generated videos, which show similar controllability to real-video training (Figs.~\ref{fig:compose_resolution}a and Fig.~\ref{fig:wan}).

\noindent\textbf{Implementation.}
Our main experiments use a frozen pretrained text-to-video DiT backbone similar to~\cite{yang2025cogvideox,zheng2024opensorademocratizingefficientvideo}; \projname~optimizes only $V^+$ directions for 300 AdamW steps and evaluates slider strengths $s_{\text{edit}}\in[0,1]$. 
We use an automatically derived attention mask at inference. We first apply the learned direction globally during the first two denoising steps, then aggregate attention from the target concept text token to visual tokens at the second step to obtain a coarse spatiotemporal soft mask that gates the remaining denoising steps. This procedure requires no user annotation. We also evaluate a mask-free variant, \textit{Ours (w/o Mask)}, which applies the learned direction globally throughout inference.

\noindent\textbf{Baselines.}
We compare with video sliders~\cite{ezra2025freesliders,chiu2026text}, and image sliders adapted to video~\cite{gandikota2024concept,gandikota2025sliderspace}.
We also include the text-driven video editing baseline, UniVideo~\cite{wei2026univideo}, by progressively strengthening the instruction (e.g., \emph{slightly} $\rightarrow$ \emph{moderately} $\rightarrow$ \emph{much} $\rightarrow$ \emph{extremely}),
and an I2I+V2V pipeline that combines the image slider~\cite{parihar2025kontinuouskontextcontinuousstrength} with propagating first-frame video edits~\cite{zi2025senorita}.

\begin{figure*}[t]
  \includegraphics[width=\textwidth]{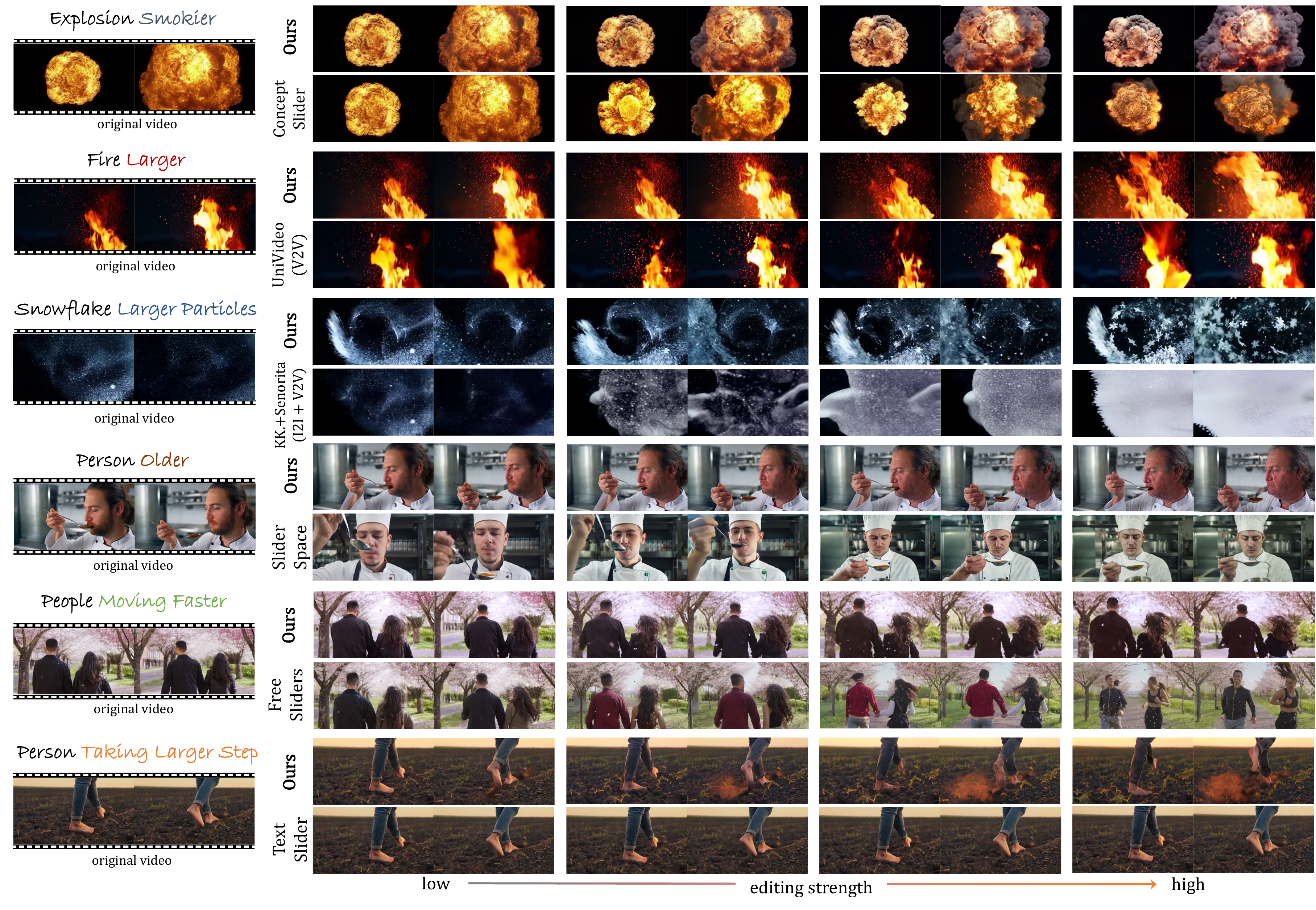}
  \vspace{-0.3cm}
  \caption{\textbf{Qualitative comparison on appearance and motion sliders.}
The first four examples show appearance sliders and the last two show motion sliders. 
\projname~yields progressive edits across strengths while better preserving identity, structure, and background.}
  \label{fig:main_results}
\end{figure*}

%% file: sections/6_new_results.tex
\section{Results}
\label{sec:results}

\subsection{Continuous Sliders for Appearance and Motion}
\label{sec:main_results}

\begin{figure*}[t]
  \centering
  \includegraphics[width=\textwidth]{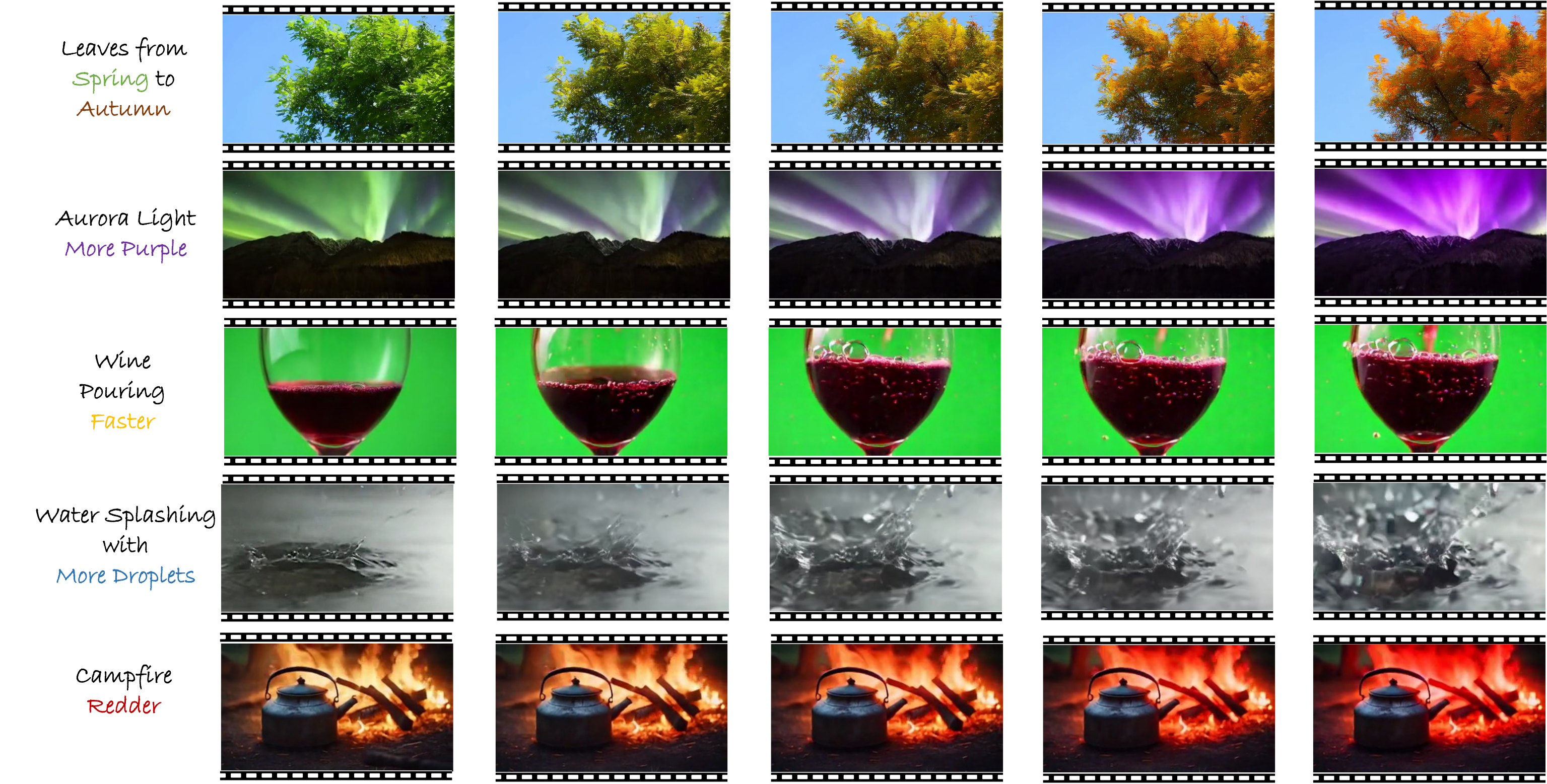}
  \caption{\textbf{Additional qualitative results}. \projname~supports diverse appearance and motion controls, including global scene-level changes and challenging partially occluded targets. Examples show continuous control over seasonal foliage and scene-wide aurora color, pouring and splash dynamics, and campfire appearance even when the target flames are partially occluded by the kettle.}
  \label{fig:more}
\end{figure*}

\noindent\textbf{Qualitative results.}
Fig.~\ref{fig:main_results} compares \projname~with representative slider and video-editing baselines on both appearance and motion attributes, with additional comparisons in the supplementary.
For appearance control, prompt-based V2V editing~\cite{wei2026univideo} changes attributes through discrete prompt variants and often yields abrupt or non-monotonic transitions.
Slider baselines such as FreeSliders~\cite{ezra2025freesliders}, Concept Slider~\cite{gandikota2024concept}, and SliderSpace~\cite{gandikota2025sliderspace} can produce visible edits, but frequently alter identity, object structure, or background as the strength increases.
More conservative methods, including Text Slider~\cite{chiu2026text}, better preserve the scene largely because the target attribute changes weakly. 
Slider-based I2I~\cite{parihar2025kontinuouskontextcontinuousstrength} followed by first-frame V2V propagation~\cite{zi2025senorita} can modify the first frame, but often fails to carry the edit through time. 
In the ``snowflake larger particles'' example, the propagated video loses the target structure and becomes washed out, showing that first-frame editing alone is insufficient for stable video sliders.
In contrast, \projname~produces progressive changes for both appearance and motion attributes while preserving the target identity and surrounding context.
Additional results in Fig.~\ref{fig:more}.

\noindent\textbf{Quantitative results: slider controllability.}
We evaluate slider behavior using the metrics of FreeSliders~\cite{ezra2025freesliders}: Conceptual Range (CR), Conceptual Smoothness (CSM), Semantic Preservation (SP), and monotonicity.
CR measures endpoint semantic span using CLIP~\cite{radford2021clip}; CSM measures the uniformity of CLIPScore changes across slider levels (lower is better); SP measures content preservation using LPIPS~\cite{zhang2018lpips}; and monotonicity measures whether consecutive CLIPScores move consistently along the attribute direction.
Since smoothness and preservation can favor methods that barely edit, we also report the overall score from~\cite{ezra2025freesliders},
$
\text{OS} = \frac{\text{CR}}{\epsilon + \text{SP}} + (1 - \text{CSM}),
$
with $\epsilon=1$. 
As shown in Table~\ref{tab:freeslider_table}, \projname~achieves the best OS, reflecting a stronger balance between edit range, smooth progression, and preservation.
The mask-free variant also outperforms all baselines in CR, monotonicity, and overall score. Adding attention-based masking further improves smoothness and preservation, yielding the best overall score.
Conservative baselines obtain competitive preservation but low range, while prompt-driven V2V obtains poor smoothness, consistent with the qualitative trends in Fig.~\ref{fig:main_results}.

\noindent\textbf{Quantitative results: VLM-based evaluation.}
Following EditVerse~\cite{ju2026editverse}, we further evaluate edit quality, identity preservation, background preservation, and temporal continuity with a VLM-based~\cite{comanici2025gemini} rubric.
Table~\ref{tab:quantitative} shows that both \projname~variants outperform the baselines across the VLM evaluation.
The mask-free variant achieves slightly higher edit quality and temporal continuity, while attention-based masking improves identity and background preservation, with comparable overall scores between the two variants. Both maintain comparable video quality and text alignment measured by PickScore~\cite{kirstain2023pick} and ViCLIP~\cite{wang2023internvid}.
These results confirm that the improvement is not only larger attribute change, but larger controllable change without sacrificing video coherence. More details are provided in the supplementary.

\input{tables/freeslider_table}

\begin{figure*}[t]
  \includegraphics[width=\textwidth]{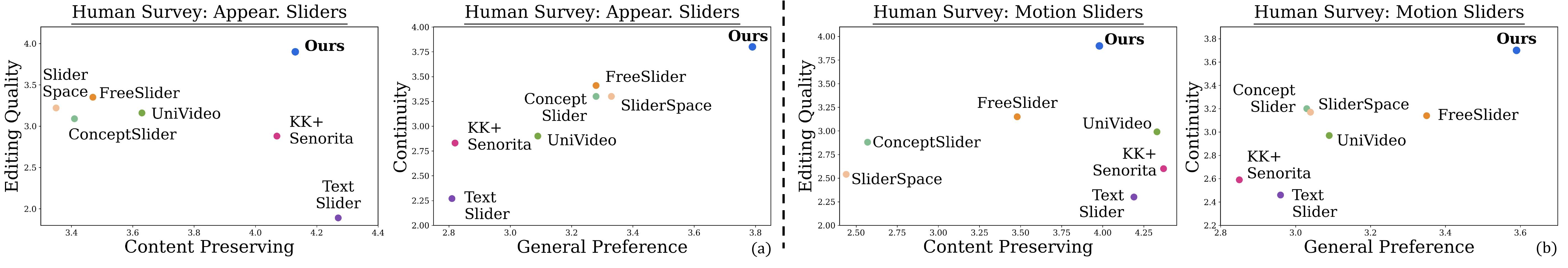}
  \vspace{-0.3cm}
  \caption{\textbf{Human survey results.} We compare methods on appearance (a) and motion sliders (b) using quadrant plots of edit quality vs.\ content preservation, and slider continuity vs.\ video general preference.}
  \label{fig:quadrant-map}
\end{figure*}

\noindent\textbf{Human evaluation.}
We conduct a human study to assess edit effectiveness and preservation, with an emphasis on motion dynamics, which are difficult to evaluate reliably with existing metrics and keyframe-based VLM scores. 
We conduct a human study on randomly sampled 32 appearance sliders and 32 motion sliders with 212 Prolific participants.
For each trial, participants view the original video and edited outputs at increasing slider strengths, then rate prompt-following/edit quality, identity/content preservation, temporal continuity, and overall preference on 0--5 Likert scales.
Each participant evaluates \projname~and all baselines in randomized order.
Fig.~\ref{fig:quadrant-map} summarizes the results as quadrant plots.
For appearance sliders, human preferences agree with the automatic metrics: \projname~achieves strong edits while preserving identity and context. For motion sliders, \projname~is clearly preferred, obtaining the highest perceived edit quality while maintaining strong preservation.
Competing methods either produce weak temporal changes or degrade coherence when pushed toward stronger motion edits.
Full protocol and additional analyses are provided in the supplementary.

\input{tables/main_table}

\input{tables/ablation_freeslider_table}

\noindent\textbf{Human--VLM agreement.}
We further examine whether the VLM-based evaluation reflects human judgments. Across the seven compared methods, VLM Overall and human general preference show strong rank agreement (Spearman $\rho=0.96$). This agreement indicates that the automatic evaluation captures the relative performance trends observed in human preference.

\begin{figure}
  \centering
  \includegraphics[width=\columnwidth]{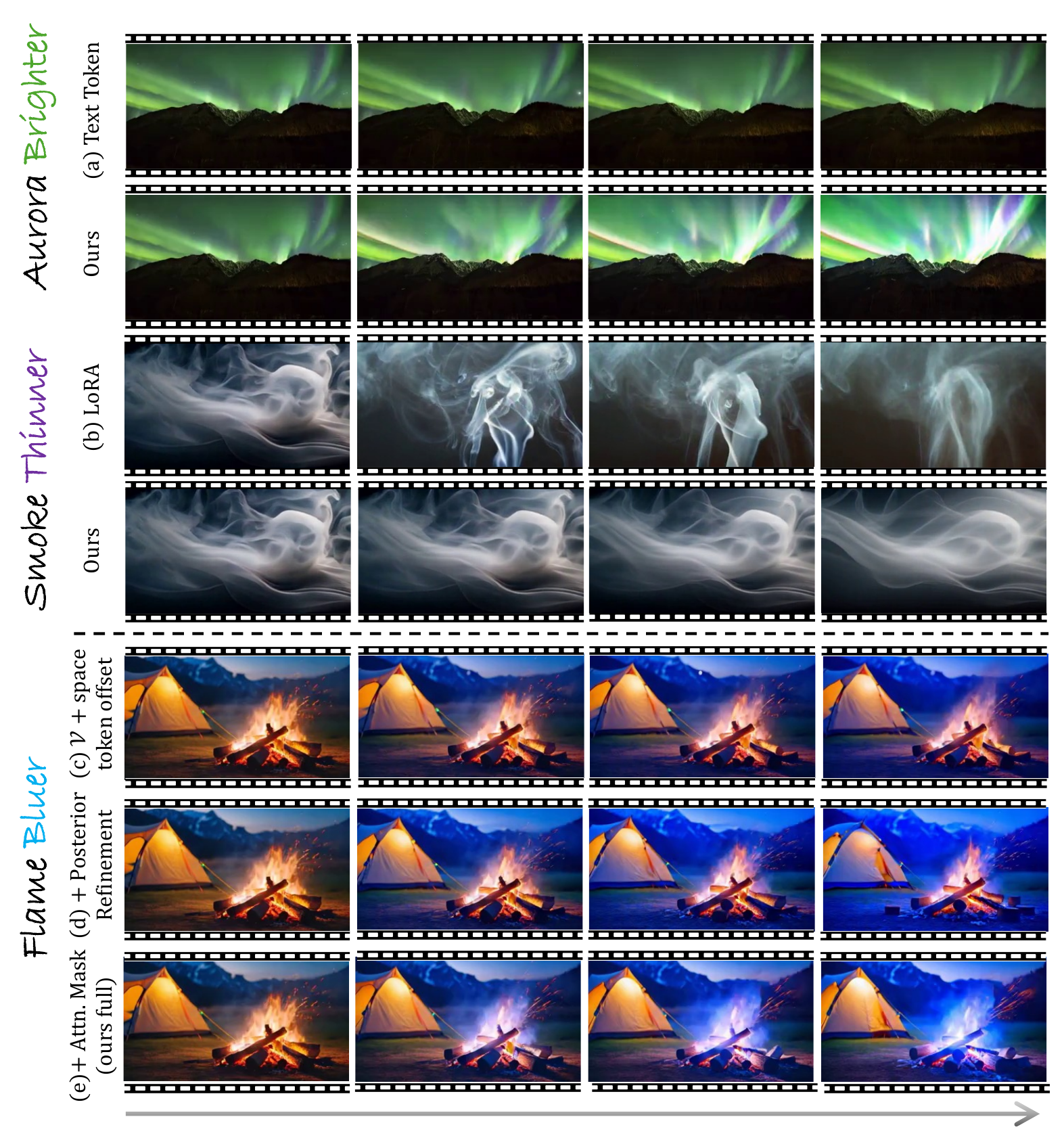}
  \vspace{-0.5cm}
  \caption{\textbf{Ablation.}
    Four increasing edit strengths are shown from left to right.
    (a)--(b) compare learned text-token and LoRA parameterizations with \projname~under the same supervision, while (c)--(e) progressively introduce the $V^+$-space token offset, posterior refinement, and attention masking.}
  \label{fig:ablation}
  \vspace{-0.5cm}
\end{figure}

\subsection{Ablation Study}
\label{sec:ablation}
\noindent\textbf{Core design choices.}
Using the same experimental setting as the main evaluation, Fig.~\ref{fig:ablation} and Table~\ref{tab:ablation} isolate the main design choices of \projname. 
We first compare alternative parameterizations of the learnable control under the same supervision. We replace the $V^+$ offset with either a learned text token, following a textual-inversion-style~\cite{gal2023an} parameterization, or LoRA-based~\cite{hu2022lora} tuning. Text-token tuning produces weak visual changes, while LoRA enables stronger edits but introduces broader changes to the background. In contrast, $V^+$ offsets directly modify visual patch tokens, providing stronger control with better background preservation.

Starting from the $V^+$ offset, we ablate the remaining components.
Posterior refinement stabilizes supervision at noisy timesteps and improves the semantic strength of the learned direction.
The mask-free variant already provides strong controllability, while attention-based masking (\textit{ours full}) confines edits to the target concept across space and time, reducing background drift and yielding the best overall score.

\noindent\textbf{Alternative supervision models.}
To isolate the effect of feature-space supervision, we conduct a separate ablation on Wan2.1-1.3B at 480p and 41 frames.
We keep all other training and evaluation settings fixed and vary only the pretrained feature encoder used for semantic direction matching, comparing InternVideo2~\cite{wang2024internvideo2}, CLIP~\cite{radford2021clip}, and PE-Core~\cite{bolya2026perception}.
We also evaluate the strongest slider baselines, SliderSpace and FreeSliders, on the same Wan2.1 backbone under matched settings.
The three supervision variants achieve VLM Overall scores of 4.122, 3.878, and 3.619, respectively, compared with 3.561 for SliderSpace and 3.211 for FreeSliders.
Although InternVideo2 performs best, \projname~maintains its advantage over both slider baselines with CLIP and PE-Core supervision, showing consistent effectiveness across different feature spaces.

\begin{figure*}[t]
  \centering
  \includegraphics[width=\textwidth]{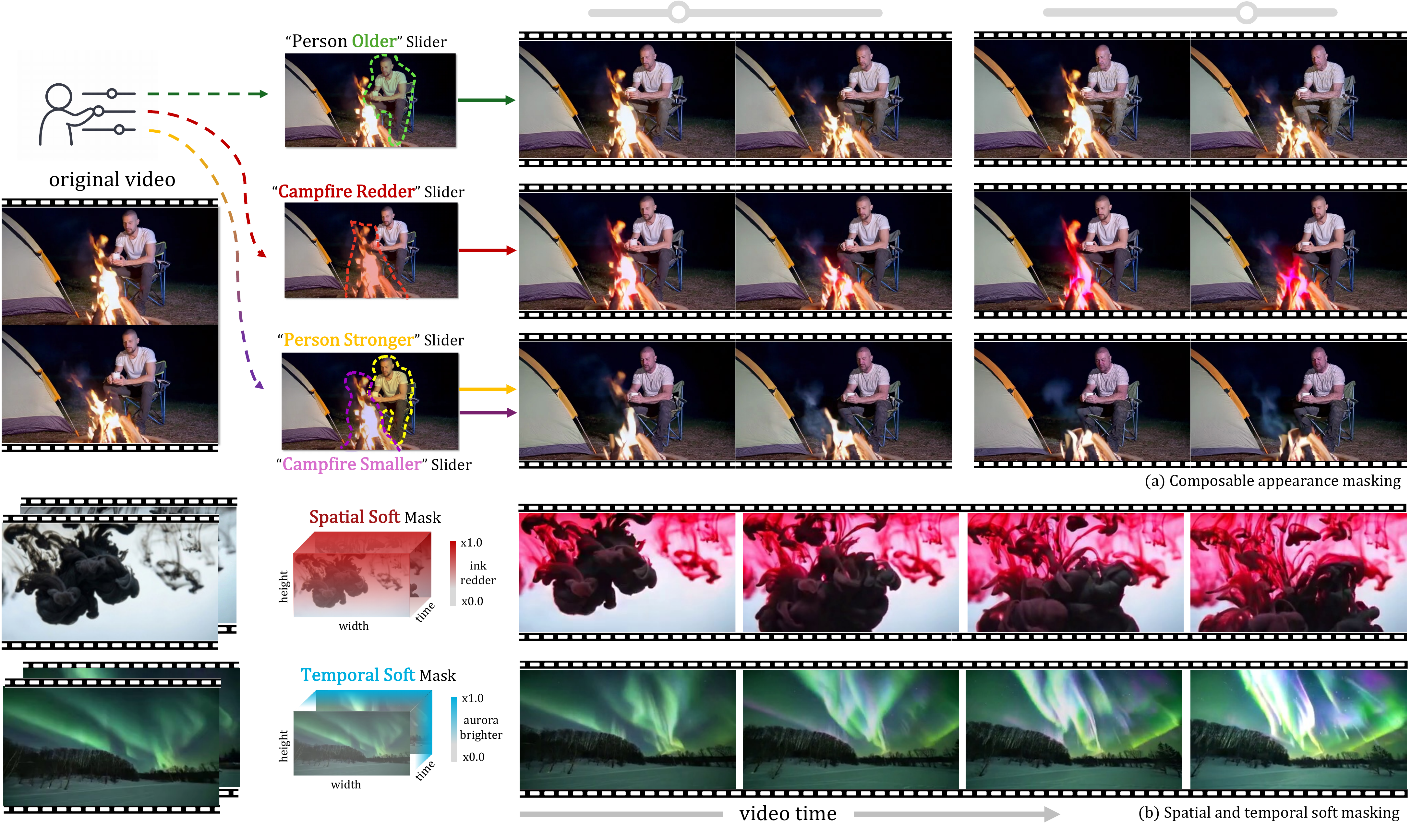}
  \vspace{-0.8cm}
  \caption{\textbf{Explicit spatiotemporal masking.} 
  (top) \emph{Composable masking:} localize different sliders to different concepts
  (person vs.\ campfire) and compose them in one video.
  (bottom) \emph{Spatial/temporal masking:} leveraging \projname~and soft masks,
  we can easily make only the top portion of the ink video redder and create a
  gradient effect, or make aurora brighter only towards the end of the video.}
  \label{fig:mask}

  \vspace{0.15cm}

  \includegraphics[width=\textwidth]{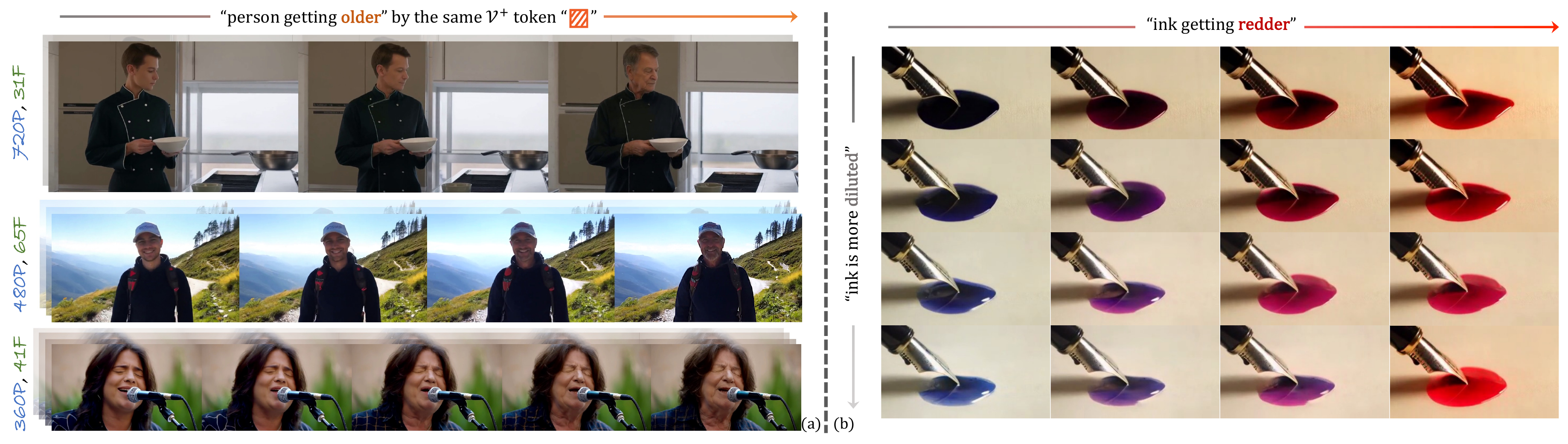}
  \vspace{-0.6cm}
  \caption{\textbf{(a)} \projname~learned direction transfers zero-shot across
  video resolutions and lengths.
  \textbf{(b)} \projname~composes attributes by combining offsets, enabling
  independent control along multiple sliders
  (e.g., ink ``redder'' and ``more diluted'').}
  \label{fig:compose_resolution}

  \vspace{0.15cm}

  \includegraphics[width=\textwidth]{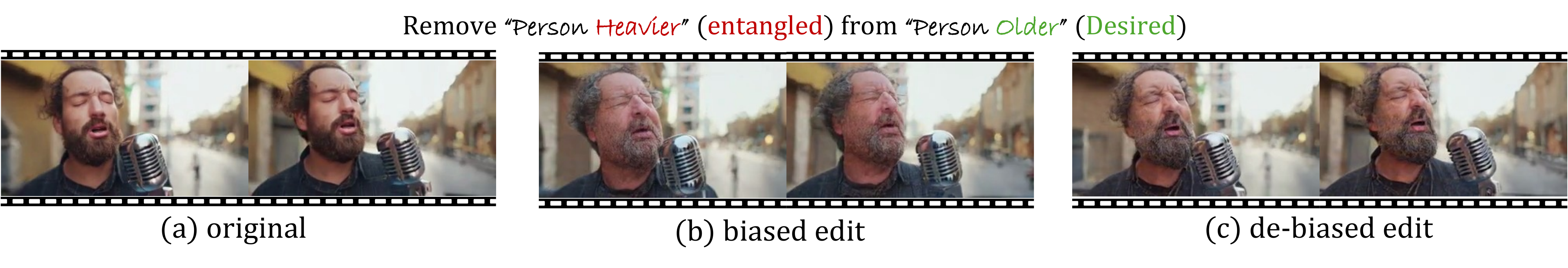}
  \vspace{-0.6cm}
  \caption{\textbf{Semantic debiasing.}
  ``Older'' edits learned from InternVideo2 can also increase body weight (b);
  debiasing removes this coupling (c).}
  \label{fig:debias}
\end{figure*}

\begin{figure*}[t]
  \centering
  \includegraphics[width=\textwidth]{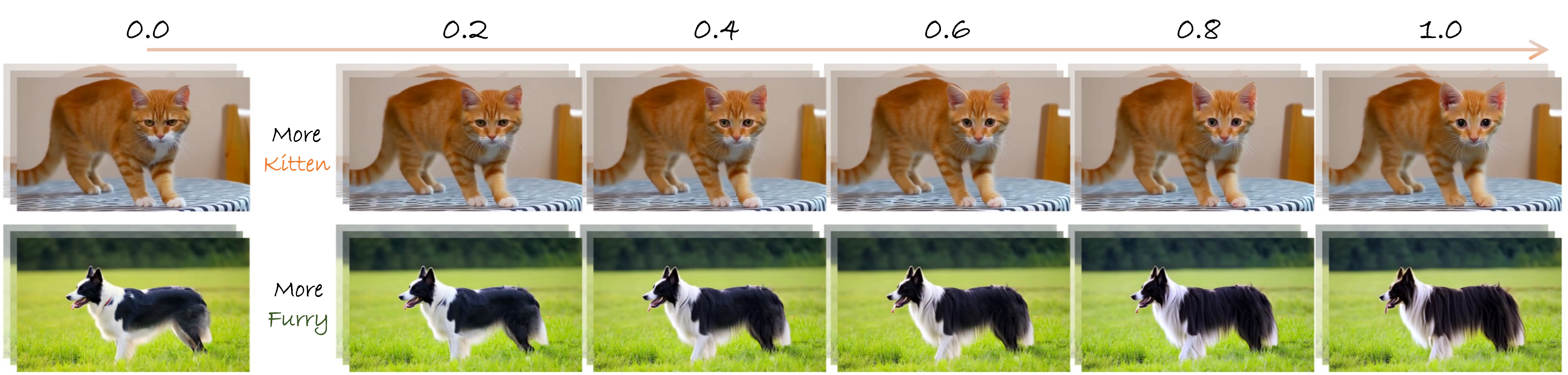}
  \vspace{-0.5cm}
  \caption{\textbf{Generalization to a different video architecture: Wan 2.1}. \projname~transfers to the Wan~2.1-T2V-1.3B backbone, enabling continuous appearance sliders by injecting offsets into Wan’s feature stream. Examples show a ``more kitten'' slider (cat) and a ``more furry'' slider (dog). }
  \label{fig:wan}

  \vspace{0.15cm}
  \includegraphics[width=\textwidth]{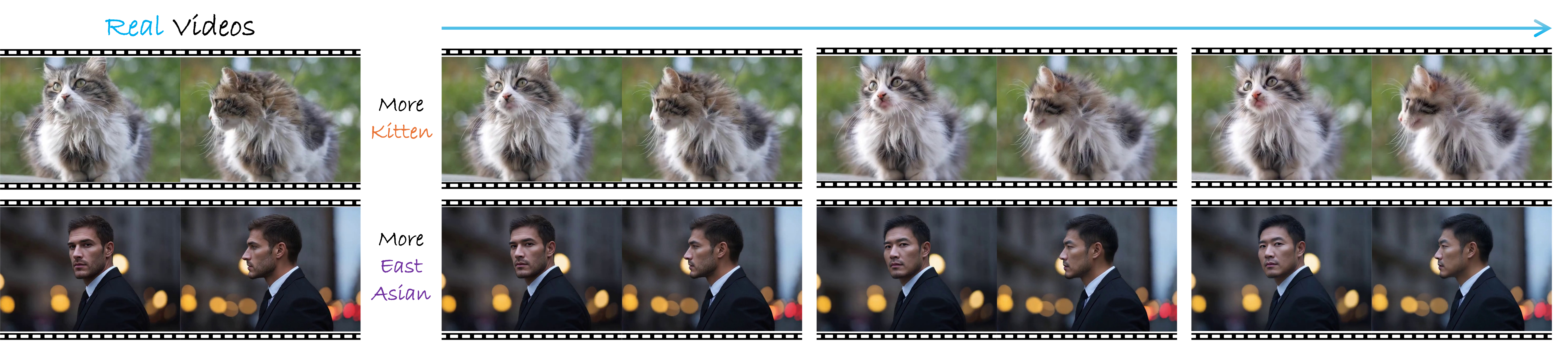}
  \vspace{-0.5cm}
  \caption{\textbf{Applying \projname~to real videos.} Learned $V^+$ directions can be directly reused for continuous video-to-video control. Increasing the dial strength produces progressive appearance changes while preserving the source video content, without additional training or per-video optimization. Source videos from~\cite{barbhuiya_redcat_vecteezy,chunaeva_man_vecteezy}.}
  \label{fig:realvideo}  
  \vspace{-0.3cm}
\end{figure*}

\subsection{Explicit Spatiotemporal Control}
\label{sec:masking}
A learned $V^+$ direction specifies \emph{what} to change, while a token mask specifies \emph{where} and \emph{when} the change applies. 
Fig.~\ref{fig:mask} demonstrates two forms of explicit spatiotemporal control. First, different masks can localize different dials to different concepts in the same video, e.g., editing the person and campfire independently without changing the rest of the scene. Second, soft masks can vary continuously over space or time: a spatial mask makes only the upper region of the ink video redder with a smooth gradient, while a temporal mask increases aurora brightness only toward the end of the video. These examples show that \projname~supports not only global sliders, but also localized and gradually varying video edits through the same learned visual dial.

\subsection{Composing Multiple Visual Dials}
\label{sec:composition}
Because $V^+$ directions are additive, multiple visual dials can be combined without retraining.
Fig.~\ref{fig:compose_resolution}(b) shows same-concept attribute composition by combining the ``ink redder'' and ``ink more diluted'' directions. The resulting two-dimensional grid provides independent control along both axes. This indicates that learned directions can remain separable under composition.

\subsection{Reusing across Prompts, Lengths, and Resolutions}
\label{sec:reuse}
Fig.~\ref{fig:compose_resolution}(a) evaluates whether a learned dial is tied to the token grid used during training. A single ``person older'' direction transfers zero-shot across different prompts, subjects, resolutions, and frame lengths, including 360P, 480P, and 720P videos. The same offset produces consistent age-related changes without retraining or changing its dimensionality. This follows from the channel-space parameterization of $V^+$: the direction can be broadcast to any supported spatiotemporal token grid of the same backbone.

\subsection{Applying Visual Dials to Real Videos}
Beyond text-to-video generation, learned $V^+$ directions can be reused for continuous editing of real videos. We combine \projname~with FlowEdit~\cite{kulikov2025flowedit}, an inversion-free editing method that evolves a source sample using the difference between source and target velocity fields. We encode the input video into the latent space and use the same text condition for the source and target branches, while injecting the learned $V^+$ direction only into the target branch. Scaling its contribution to the target velocity field provides continuous video-to-video control without retraining the dial or optimizing for the input video. As shown in Fig.~\ref{fig:realvideo}, the learned directions produce progressive appearance changes on real videos while preserving the source structure.

\subsection{Generalizing across Video Backbones}
\label{sec:backbones}
We also test whether the token-offset principle is specific to our main DiT backbone. In addition to the T2V backbone used in the main experiments, we instantiate \projname~on Wan~2.1~\cite{wan2025wanopenadvancedlargescale} by injecting offsets into its visual patch token stream and learning $V^+$ directions with the same objective. As shown in Fig.~\ref{fig:wan}, the learned directions produce smooth strength-dependent edits on Wan. This suggests that \projname~provides a general control mechanism beyond a single video generator.

\noindent\textbf{Training efficiency and scalability.}
On Wan2.1-1.3B at 480p with 9 feature keyframes, optimizing one visual dial takes 12 minutes on a single H100 GPU (80~GB) with 58~GB peak memory. On Wan2.1-14B, the same optimization takes 18 minutes on two H100 GPUs (80~GB each), with 44/42~GB peak memory per GPU.

\subsection{Debiasing Attribute Directions}
\label{sec:debiasing}
Since \projname~learns from pretrained understanding models, a direction can inherit entanglements from the supervision space.
Fig.~\ref{fig:debias} shows an ``older'' direction that also increases body weight, reflecting a spurious age--body-shape correlation.
We mitigate this by projecting out biased principal directions before training, following a debiasing strategy similar to~\cite{tanjim2024mitigatebias}.
After debiasing, the same ``older'' edit better preserves body shape while still increasing perceived age (Fig.~\ref{fig:debias}c).
The \projname~learned direction reflects its supervision signal, but its compact explicit form allows for inspection and modification before generation.

%% file: tables/freeslider_table.tex
\begin{table}[t]
\footnotesize
  \centering
  \caption{Quantitative evaluation of slider controllability.}
  \label{tab:freeslider_table}
  \setlength{\tabcolsep}{5pt}
  \renewcommand{\arraystretch}{0.95}
  \resizebox{\linewidth}{!}{
  \begin{tabular}{lcccc|c}
    \toprule
     
    & \multicolumn{3}{c}{\textbf{Editing Quality}} 
    & \multicolumn{1}{c|}{\textbf{Preserve}}
    & \multicolumn{1}{c}{} \\
    
    \cmidrule(lr){2-4}
    \cmidrule(lr){5-5}
    \textbf{Method}
    & \textbf{Range(CR)} $\uparrow$
    & \textbf{Mono.} $\uparrow$
    & \textbf{Smooth(CSM)} $\downarrow$
    & \textbf{LPIPS(SP)} $\downarrow$
    & \textbf{Overall} $\uparrow$\\
    
    \midrule
    Concept Slider & 0.210 & 0.512 & 0.735 & 0.391 & 0.416 \\
    SliderSpace   & 0.451 & 0.550 & 0.676 & 0.403 & 0.645 \\
    FreeSliders    & 0.378 & 0.567 & 0.537 & 0.296 & 0.755 \\
    Text Slider    & 0.004 & 0.501 & \textbf{0.262} & \cellcolor{gray!20}0.110 & 0.742 \\
    \midrule
    Slider-based I2I+V2V & 0.196 & 0.542 & \cellcolor{gray!20}0.373 & \textbf{0.085} & 0.808 \\
    Text-based V2V       & 0.276 & 0.534 & 1.154 & 0.440 & 0.037 \\
    \midrule
    Ours (w/o Mask)       & \textbf{0.499} & \cellcolor{gray!20}0.570 & 0.440 & 0.298 & \cellcolor{gray!20}0.944 \\
    \textbf{Ours}          & \cellcolor{gray!20}{0.460} & \textbf{0.573} & 0.386 & 0.249 & \textbf{0.982} \\
    \bottomrule
  \end{tabular}
  }
\end{table}

%% file: tables/main_table.tex
\begin{table}[t]
  \centering
  \caption{Quantitative results on VLM and video quality metrics.}
  \label{tab:quantitative}
  \resizebox{\linewidth}{!}{%
    \begin{tabular}{lccccccc}
      \toprule
      \textbf{Method} 
      & \multicolumn{5}{c}{\textbf{VLM Evaluation}} 
      & \textbf{Video Qual.} 
      & \textbf{Text Align.} \\
      
      \cmidrule(lr){2-6}
      \cmidrule(lr){7-7}
      \cmidrule(lr){8-8}
      
      & \textbf{Edit Qual.} $\uparrow$
      & \textbf{ID Pre.} $\uparrow$
      & \textbf{BG Pre.} $\uparrow$
      & \textbf{Continuity} $\uparrow$
      & \textbf{Overall} $\uparrow$
      & \textbf{PickScore} $\uparrow$
      & \textbf{ViCLIP} $\uparrow$ \\
      
      \midrule
      Concept Slider   & 3.635 & 4.722 & 4.599 & 3.654 & 4.153 & 19.610 & 24.737 \\
      SliderSpace     & 3.743 & 4.722 & 4.603 & 3.738 & 4.202 & 19.659 & 24.813 \\
      FreeSliders      & 3.932 & 4.830 & 4.775 & 3.936 & 4.368 & \cellcolor{gray!20}19.697 & \textbf{25.242} \\
      Text Slider      & 2.771 & \cellcolor{gray!20}4.970 & \textbf{4.974} & 2.833 & 3.887 & \textbf{19.726} & 24.936 \\
      \midrule
      \makecell{Kontinuous Kontext\\+Senorita (I2I+V2V)} & 2.578 & 4.868 & 4.940 & 2.518 & 3.726 & 19.567 & 24.353 \\
      UniVideo (V2V)  & 3.032 & 4.754 & 4.874 & 2.786 & 3.862 & 19.461 & 23.722 \\
      \midrule
      Ours (w/o Mask) & \textbf{4.224} & 4.946 & 4.905 & \textbf{4.284} & \textbf{4.590} & 19.629 & \cellcolor{gray!20}25.005 \\
      Ours            & \cellcolor{gray!20}4.165 & \textbf{4.988} & \cellcolor{gray!20}4.959 & \cellcolor{gray!20}4.234 & \cellcolor{gray!20}4.587 & 19.651 & 24.942 \\
      \bottomrule
    \end{tabular}%
  }
\end{table}

%% file: tables/ablation_freeslider_table.tex
\begin{table}[t]
\footnotesize
  \centering
  \caption{Ablation analysis of components for controllable editing.}
  \label{tab:ablation}
  \setlength{\tabcolsep}{5pt}
  \renewcommand{\arraystretch}{0.95}
  \resizebox{0.95\linewidth}{!}{
  \begin{tabular}{lccccc}
    \toprule
     
    & \multicolumn{3}{c}{\textbf{Editing Quality}} 
    & \multicolumn{1}{c}{\textbf{Preserve}} 
    & \multicolumn{1}{c}{} \\
    
    \cmidrule(lr){2-4}
    \cmidrule(lr){5-5}
    \textbf{Method}
    & \textbf{Range} $\uparrow$
    & \textbf{Mono.} $\uparrow$
    & \textbf{Smooth} $\downarrow$
    & \textbf{LPIPS} $\downarrow$
    & \textbf{Overall} $\uparrow$\\
    
    \midrule
    Text Token   & 0.011 & 0.458 & \textbf{0.241} & \textbf{0.129} & 0.769\\
    LoRA         & 0.393 & 0.550 & 0.758 & 0.434 & 0.516\\
    \midrule
    $\mathcal{V}^{+}$-Space Offset   & 0.412 & 0.566 & 0.419 & 0.270 & 0.905\\
    + Posterior Refine                & \textbf{0.499} & \cellcolor{gray!20}0.570 & 0.440 & 0.298 & \cellcolor{gray!20}0.944\\
    + Attn.\ Mask (ours full)        & \cellcolor{gray!20}0.460 & \textbf{0.573} & \cellcolor{gray!20}0.386 & \cellcolor{gray!20}0.249 & \textbf{0.982}\\
    \bottomrule
  \end{tabular}
  }
\end{table}

%% file: sections/7_conclusion.tex
\section{Conclusion}
We present \projname, which introduces continuous slider control to pretrained text-to-video models by learning additive directions in a compact visual dial space $V^+$ over patch-token channels. The learned directions are semantic: they yield smooth, predictable changes for both appearance and motion, and support spatiotemporal localization, composition, and reuse across prompts, resolutions, and video lengths, and can be applied to real videos without retraining the generator; \projname~further generalizes across video backbones. We hope this opens further work on interactive, fine-grained video controllability.

\noindent\textbf{Limitations.} \projname~relies on pretrained video understanding models (e.g., InternVideo2) for appearance supervision, and thus inherits the entanglements and biases of that space. Beyond high-level correlations (e.g., age with body weight), low-level attributes such as color can be coupled with other visual factors, where simple subspace-projection debiasing may be insufficient to isolate.

%% file: supplimentary/sections/x_suppl_new.tex

\section*{Appendix}

In this supplementary material, we provide additional evaluation protocols,
experimental details, ablations, implementation details, and qualitative
results that complement the main paper.


\tableofcontents


\section{Evaluation Details}
\label{sec:evaluation_details}


\subsection{Human Evaluation Study}
\label{sec:human_study}

\begin{figure*}[t]
    \centering
    \includegraphics[width=0.92\textwidth]
    {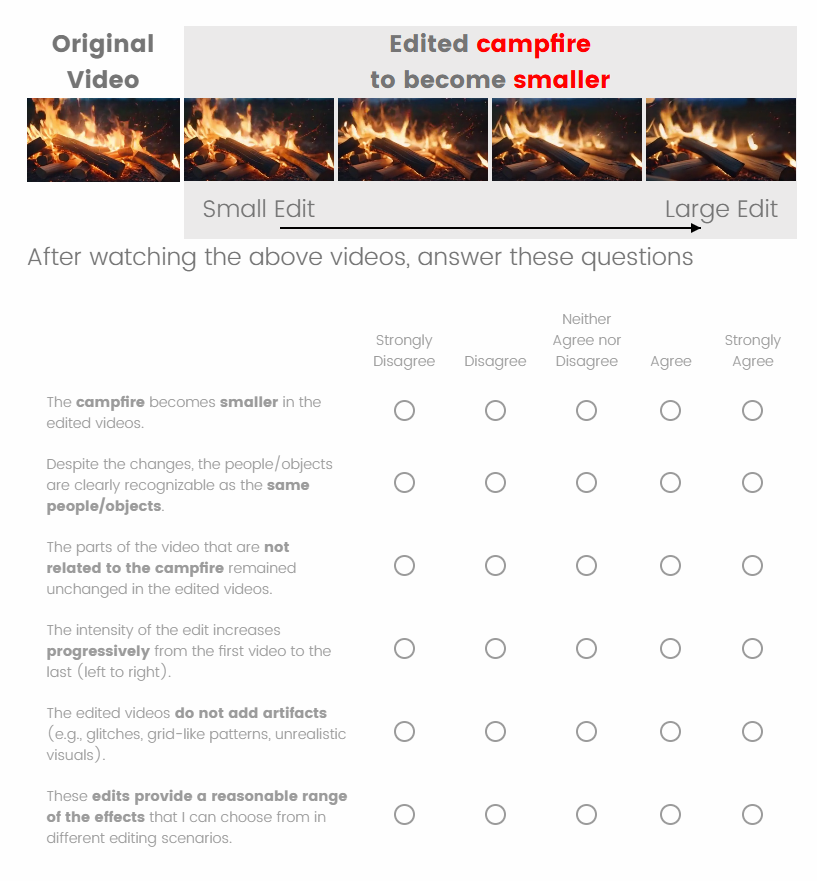}
    \caption{
        A screenshot example from our human evaluation study.
    }
    \label{fig:survey_screenshot}
\end{figure*}

We conducted a human evaluation study using a Qualtrics survey distributed
through Prolific. As shown in Fig.~\ref{fig:survey_screenshot}, each survey
question presented an original video together with four edited videos.
The edited videos were generated by one of six baseline methods or by our
method, chosen randomly, with edit strength increasing progressively across
the four edits.

The baselines included Concept Slider~\cite{gandikota2024concept},
SliderSpace~\cite{gandikota2025sliderspace},
FreeSliders~\cite{ezra2025freesliders},
Text Slider~\cite{chiu2026text},
Kontinuous Kontext~\cite{parihar2025kontinuouskontextcontinuousstrength}
combined with Senorita~\cite{zi2025senorita} as a slider-based
image-to-image (I2I) plus first-frame propagation video editing baseline,
and UniVideo~\cite{wei2026univideo} as a text-based video-to-video (V2V)
editing baseline.

Below the videos, participants rated six statements on a Likert scale.
These questions were aligned with our VLM-based evaluation and were designed
to measure (1) edit quality and progression, (2) content preservation, and
(3) subjective human preference.
On each survey page, participants were shown five sets of video edits, one
from each method being compared, together with the corresponding questions.
This design ensured that every participant evaluated all compared baselines
the same number of times, enabling a fair comparison across methods.
The order of the baselines was randomized.

\begin{figure}
    \centering
    \includegraphics[width=0.9\columnwidth]
    {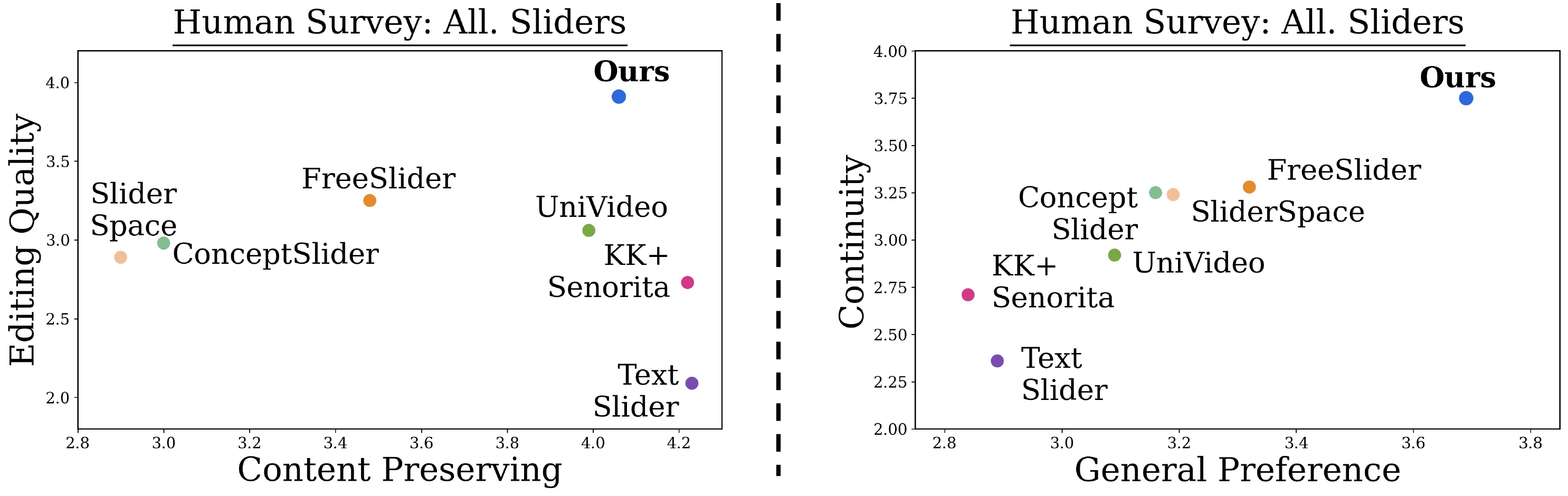}
    \caption{
        Human evaluation across appearance and motion sliders.
        We compare edit quality against content preservation, and slider
        continuity against general preference.
        Results aggregate appearance and motion sliders.
    }
    \label{fig:human_quadrants}
\end{figure}

We generated a total of 64 editing directions for this study,
corresponding to real-world video editing tasks such as making a person
walk faster or making a fire brighter.
Among them, 32 editing directions targeted appearance attributes, while
the other 32 targeted temporal or dynamic changes.
As shown in Fig.~\ref{fig:survey_screenshot}, the editing prompt was inserted
directly into the survey; for example,
\textit{campfire/smaller} could be replaced with \textit{speed/slower}.

In total, we recruited \textbf{212 participants} through Prolific.
We first compared \projname~against video-slider-based methods.
Specifically, FreeSliders~\cite{ezra2025freesliders} and
Text Slider~\cite{chiu2026text}
are originally designed for video slider control, while
Concept Slider~\cite{gandikota2024concept} and
SliderSpace~\cite{gandikota2025sliderspace}
are image-slider methods that we adapted to our base text-to-video (T2V)
generation model to obtain video-slider baselines.

For this comparison, we recruited \textbf{122 participants}.
Each participant rated 20 sets of video edits, corresponding to
4 different editing instructions evaluated across 5 methods
(the four video-slider-based baselines and ours).

We recruited another \textbf{90 participants} to evaluate the slider-based
I2I+V2V and text-based V2V baselines.
Each participant rated 12 sets of video edits, corresponding to
4 different editing instructions evaluated across 3 methods:
Kontinuous Kontext~\cite{parihar2025kontinuouskontextcontinuousstrength}
combined with Senorita~\cite{zi2025senorita},
UniVideo~\cite{wei2026univideo}, and our method.

\begin{table}[t]
    \centering
    \footnotesize
    \caption{Mapping our evaluation metrics to the Likert statements in our study.}
    \begin{tabular}{p{0.18\linewidth}p{0.73\linewidth}}
        \toprule
        Metric & Question \\
        \midrule

        Editing Quality &
        The \textbf{attribute} becomes \textbf{modifier} in the edited videos. \\

        ID Preserve &
        Despite the changes, the people/objects are clearly recognizable
        as the \textbf{same people/objects}. \\

        BG Preserve &
        The parts of the video that are \textbf{not related to the attribute}
        remained unchanged in the edited videos. \\

        Continuity &
        The intensity of the edit increases \textbf{progressively}
        from the first video to the last (left to right). \\

        Artifacts &
        The edited videos \textbf{do not add artifacts}
        (e.g., glitches, grid-like patterns, unrealistic visuals). \\

        Useful &
        These \textbf{edits provide a reasonable range of the effects}
        that I can choose from in different editing scenarios. \\

        \bottomrule
    \end{tabular}
    \label{tab:questions}
\end{table}

\begin{figure}
    \centering
    \includegraphics[
        width=0.96\columnwidth,
        keepaspectratio
    ]{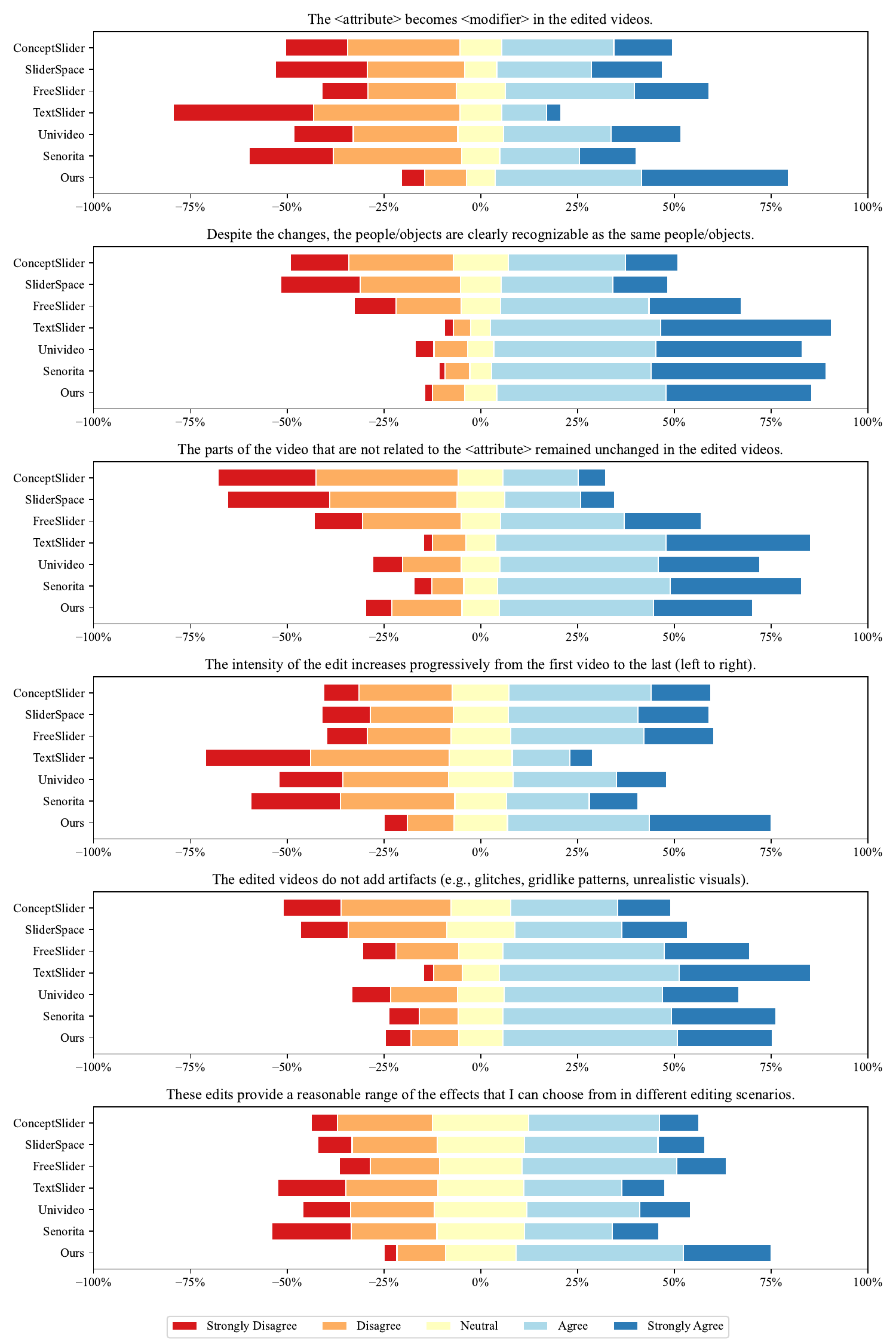}
    \caption{
        Distribution of Likert ratings in our human evaluation study.
    }
    \label{fig:likert-barchart}
\end{figure}

We report the average Likert-scale ratings in
Table~\ref{tab:suppl_human_eval}, where ``strongly agree''
corresponds to 5 and ``strongly disagree'' corresponds to 1.
Participants generally agreed that our method can produce progressive edits
while preserving subject identity and background consistency.
These results largely follow the same trends as the VLM-based evaluation,
although the human ratings exhibit substantially greater variance,
particularly for identity (ID) and background (BG) preservation.

Participants also agreed more strongly than for other baselines that the
``edits provide a reasonable range of the effects that I can choose from in
different editing scenarios,'' as reflected in the
``General Preference'' column of Table~\ref{tab:suppl_human_eval}.
This suggests a clear subjective preference for our proposed editing approach.

We further visualize the Likert-scale distributions in
Fig.~\ref{fig:likert-barchart}.
Each segment of the diverging stacked bar chart represents the percentage
of participants who assigned a particular Likert rating to a given question
for a given baseline.
Our method consistently receives more ``strongly agree'' and ``agree''
responses across most questions, all of which were phrased such that
agreement indicates a positive evaluation.

For content-preservation-related questions,
Text Slider and Kontinuous Kontext + Senorita sometimes receive comparable
or even higher agreement.
We attribute this to the fact that these methods tend to make minimal changes,
which also explains their many ``strongly disagree'' responses on
edit-quality-related questions in Fig.~\ref{fig:likert-barchart}.


\input{supplimentary/tables/human_survey}

\FloatBarrier


\subsection{VLM Evaluation Prompt}
\label{sec:vlm_prompt}

For VLM-based evaluation, we jointly evaluate the complete slider trajectory
using five frames sampled from the same timestamp, corresponding to the
original video and four progressively stronger edit strengths.
The evaluator considers prompt following, identity preservation, background
consistency, and progressive edit intensity.

We use the following prompt template.

\begin{lstlisting}[style=vlmprompt]
You are a meticulous video editing quality evaluator.

Your task is to assess a VIDEO EDIT SLIDER by analyzing 5 images sampled from the SAME timestamp,
corresponding to 5 increasing slider scales.

Scale definition:
- Scale 1: Original video frame (before editing)
- Scale 2-5: Edited frames with progressively stronger edit strength

Concept: "{concept}"
Target attribute direction: "{attribute}"
Goal: Make the concept become more "{attribute}" progressively from Scale 2 to Scale 5.

Instructions:
Analyze the 5-scale sequence as a whole and evaluate how well the editing slider satisfies the editing goal
while preserving visual consistency.
- All 5 images depict the SAME moment in time.
- The main subject identity and background should remain the SAME across all scales despite the edit.
- Stronger edits NEVER justify replacing the subject or changing the background.

Score the ENTIRE 5-scale sequence on 4 criteria, each integer in [0,5] where higher is better.
Provide a short justification for each score.

You will evaluate the slider across FOUR criteria.
For each criterion, provide a score from 0 (worst) to 5 (best) and a brief justification.

1. Prompt Following (Score: 0-5)

Question:
Does the slider edit direction correctly and consistently make the SAME "{concept}"
appear more "{attribute}" from Scale 2 to Scale 5?

Scoring Guide:
- 5: The edit perfectly follows the prompt, with Scale 5 being the strongest and most aligned.
- 4: The edit mostly follows the prompt with minor weaknesses.
- 3: The edit partially follows the prompt but is ambiguous or inconsistent.
- 2: The edit weakly reflects the prompt.
- 1: The edit barely relates to the prompt.
- 0: The prompt is ignored or contradicted.

2. Identity Preserving (Score: 0-5)

Question:
Does the main subject remain the SAME identity and category across all scales?

If the subject is replaced, changes category, or becomes a different person/object at ANY scale,
this is considered a severe violation and will be heavily penalized.

Scoring Guide:
- 5: Identity is perfectly preserved across all scales.
- 4: Very minor visual changes but clearly the same subject.
- 3: Noticeable drift, but identity is still mostly recognizable.
- 2: Major identity inconsistency or partial replacement.
- 1: Severe identity change.
- 0: The subject is completely replaced or unrecognizable.

3. Background Consistency (Score: 0-5)

Question:
Have the regions that should NOT be edited (background, scene layout, camera viewpoint)
remained stable across all scales?

Scoring Guide:
- 5: Background is perfectly preserved and stable.
- 4: Minor, subtle background changes.
- 3: Noticeable but non-catastrophic background drift.
- 2: Significant background changes or redraw.
- 1: Severe background inconsistency.
- 0: Background is completely altered.

4. Progressive Intensity (Score: 0-5)

Question:
Does the edit strength increase monotonically and smoothly from Scale 1 to Scale 5?

Scoring Guide:
- 5: Smooth, monotonic increase with clear ordering from weak to strong.
- 4: Mostly monotonic with minor irregularities.
- 3: Inconsistent progression.
- 2: Weak or unclear progression.
- 1: Reversed or chaotic progression.
- 0: No meaningful progression.

Return STRICT JSON ONLY (no markdown) with this schema:
{
  "prompt_following": {"score": <int 0-5>, "reason": <string>},
  "id_preserving": {"score": <int 0-5>, "reason": <string>},
  "background_consistency": {"score": <int 0-5>, "reason": <string>},
  "progressive_intensity": {"score": <int 0-5>, "reason": <string>}
}
\end{lstlisting}

\FloatBarrier


\section{Experimental Setup and Baselines}
\label{sec:experiment_setup}


\subsection{Concepts, Attributes, and Slider Tasks}

We study slider-style control defined over a concept--attribute pair.
A \textit{concept} is the main entity or phenomenon described in the prompt
(e.g., campfire, ink, person) with localized spatiotemporal support in the
video.

An \textit{attribute} is a continuous scalar property of that concept that
can be modulated while preserving scene layout and identity
(e.g., color intensity, size, dilution, motion magnitude).

A \textit{slider task} evaluates whether a method can produce progressive,
monotonic changes along the attribute scale under the same prompt and seed.

We consider two families of tasks:
appearance sliders modulate object-centric spatial attributes under strict
localization, with background and identity preserved;
motion sliders modulate motion magnitude while preserving the underlying
motion pattern.


\subsection{\projname~Dataset and Evaluation Protocol}

We evaluate on 12 concepts spanning particle systems, volumetric phenomena,
fluids, and articulated subjects, with 5 attributes per concept.
Appearance offsets are trained on small, concept-specific video--text
collections without paired edits.
Motion offsets are trained on a few hundred green-screen clips and applied
universally across prompts during inference.

For each concept--attribute pair, we generate 16 base videos and evaluate
5 slider strengths under identical prompts, seeds, and inference budgets,
resulting in roughly 4,500 videos per method.


\subsection{Implementation Details}

We optimize token offsets on a frozen pretrained text-to-video DiT model,
sharing a similar architecture to
~\cite{yang2025cogvideox,zheng2024opensorademocratizingefficientvideo}.

Unless specified otherwise, we use
$\lambda_a{=}0.5$ for appearance and
$\lambda_m{=}5.0$ for motion,
and train offsets for 300 steps with AdamW
(lr $=10^{-5}$).

Training uses 32-frame videos at 24 FPS with $320\times176$ resolution;
at inference we broadcast offsets to longer sequences (up to 64 frames).
We use classifier-free guidance with text scale
$s_{\text{txt}}{=}4.5$
and vary the edit strength with
$s_{\text{edit}}\in[0,1]$
across all experiments.


\subsection{Baselines}

We compare against representative methods for continuous control.
Our primary video slider baselines are
FreeSliders~\cite{ezra2025freesliders}
and Text Slider~\cite{chiu2026text}.

We also adapt image-domain sliders,
Concept Sliders~\cite{gandikota2024concept}
and SliderSpace~\cite{gandikota2025sliderspace},
to videos by applying their learned directions consistently across frames.

Finally, we include a strong text-driven video editing baseline,
UniVideo~\cite{wei2026univideo},
by progressively strengthening the instruction
(e.g., \emph{slightly} $\rightarrow$ \emph{moderately}
$\rightarrow$ \emph{much} $\rightarrow$ \emph{extremely}),
and an I2I+V2V pipeline that combines the image slider
Kontinuous Kontext~\cite{parihar2025kontinuouskontextcontinuousstrength}
with Senorita~\cite{zi2025senorita}
for propagating first-frame edits.


\subsection{Details of Slider-Based I2I-V2V and Text-Based V2V}
\label{sec:i2v_v2v}

We first generate a base video from our underlying video generation model
using the same prompt and same random seed.
For the slider-based I2I baseline, we use
Kontinuous Kontext~\cite{parihar2025kontinuouskontextcontinuousstrength},
a slider-based image editing pipeline.
Given an editing instruction, Kontinuous Kontext produces the corresponding
edited image at different strength levels.

We use the first frame of the generated video as the image to be edited.
After obtaining the edited first frames, we apply a first-frame
propagation-based video editing model,
Senorita~\cite{zi2025senorita}.
Given the edited first frame and the original video, Senorita propagates
the first-frame edit to the entire video, producing the final edited videos.

For the text-based V2V baseline, we use
UniVideo~\cite{wei2026univideo},
a text-driven video-to-video editing model.
Given the original video and an editing instruction, we construct a set of
modified editing instructions that explicitly encode different edit strengths
(e.g., \emph{slightly} $\rightarrow$ \emph{moderately}
$\rightarrow$ \emph{much} $\rightarrow$ \emph{extremely}).

For example, if the original instruction is
``make the campfire redder,''
the corresponding scale-specific instructions are:

\begin{itemize}
    \item ``make the campfire \emph{slightly} redder''
    \item ``make the campfire \emph{moderately} redder''
    \item ``make the campfire \emph{much} redder''
    \item ``make the campfire \emph{extremely} redder''
\end{itemize}

We use these prompts as inputs to UniVideo and perform each edit independently.

\FloatBarrier


\section{Dataset Details}
\label{sec:dataset}


\subsection{Training Dataset}

\begin{table}[t]
    \centering
    \footnotesize
    \caption{Concepts and attributes used for evaluation.}
    \begin{tabular}{p{0.18\linewidth}p{0.73\linewidth}}
        \toprule
        Concepts & Attributes \\
        \midrule

        aurora &
        brighter, dimmer, larger, more purple, smaller \\

        bubble &
        more likely to burst quickly, denser, larger, smaller, sparser \\

        campfire &
        bluer, brighter, larger, smaller, warmer \\

        confetti &
        more chaotic, more color-saturated, denser, sparser, more stable \\

        explosion &
        brighter, dimmer, larger, smaller, more smoky \\

        ink &
        more diluted, greener, redder, more spread out, thicker \\

        person &
        more curly hair, happier, heavier, older, from walk to run \\

        smoke &
        calmer, darker, thicker, thinner, more turbulent \\

        snowflake &
        brighter, denser, larger particle, smaller particle, sparser \\

        spark &
        brighter, more chaotic, denser, sparser, more stable \\

        water splash &
        more chaotic, less droplet, more droplet, more stable,
        higher viscosity \\

        speed &
        higher motion magnitude, lower motion magnitude \\

        \bottomrule
    \end{tabular}
    \label{tab:concept_attribute}
\end{table}

Table~\ref{tab:concept_attribute} summarizes the concepts and attributes
used in our evaluation.
For each concept, we construct a small internal text--video paired dataset
containing a few hundred training samples with minimal supervision.

We first use keyword-based filtering to select text--video pairs whose text
descriptions explicitly mention the target concept, followed by light manual
cleaning and deduplication.
This produces training pairs where the target concept is named in the prompt
and usually visually centered in the corresponding video.

For example, for the \emph{campfire} concept, one training sample uses the
text prompt:
``A campfire in an abandoned city street at night, with urban ruins
surrounding it.''

For the \emph{speed} concept, we additionally curate a few hundred
green-screen training videos.
The motivation for using the green-screen videos is to reduce background
noise and focus the supervision more directly on foreground motion changes.
Although these green-screen videos are used during training, empirically we
found the learned model generalizes well to natural scenes at inference time.


\subsection{Test-Time Settings}

At test time, for each concept--attribute pair shown in
Table~\ref{tab:concept_attribute},
we construct 16 base prompts that are not included in the training set.
Each prompt explicitly contains the target concept.

For each prompt, we generate two base videos using different random seeds.
For every concept--attribute--prompt--seed combination, we then generate
edited videos at scales 1--4, where scale 0 corresponds to the original
video.

This evaluation protocol enables us to assess controllable editing across
diverse concepts, prompts, random seeds, and edit strengths.



\section{Additional Ablations}
\label{sec:additional_ablation}


\subsection{Choice of Learnable Module}
\label{sec:learnable_module}

To isolate the effect of the learnable representation, we additionally
consider two learned alternatives under the same supervision:
LoRA-based tuning and a learned text-token baseline, where the latter
optimizes an inserted text token in the style of textual inversion or
P-Tuning.

These variants change the learnable module while keeping the supervision
and evaluation protocol unchanged, allowing us to directly compare parameter-space, text-token-space, and visual-token-space control.




\begin{table*}[t]
    \centering
    \small
    \caption{
        \textbf{Transformer layer ablation.}
        We compare different strategies for distributing the learned
        visual-dial directions across the 30 DiT layers of Wan~2.1-1.3B.
    }
    \label{tab:layer_ablation}

    \begin{tabular}{lccccc}
        \toprule
        \textbf{Configuration}
        & \textbf{Range (CR)} $\uparrow$
        & \textbf{Mono.} $\uparrow$
        & \textbf{Smooth. (CSM)} $\downarrow$
        & \textbf{LPIPS (SP)} $\downarrow$
        & \textbf{Overall} $\uparrow$ \\
        \midrule

        Layer-specific, all layers
        & \textbf{0.5073}
        & \textbf{0.5917}
        & 0.2517
        & 0.0940
        & \textbf{1.2120} \\

        Shared, all layers
        & 0.1933
        & 0.5750
        & 0.2425
        & 0.0652
        & 0.9390 \\

        Early layers only
        & 0.1017
        & 0.5667
        & \textbf{0.2035}
        & 0.0575
        & 0.8927 \\

        Middle layers only
        & 0.4714
        & 0.5833
        & 0.2411
        & 0.0715
        & 1.1988 \\

        Late layers only
        & 0.0288
        & 0.5167
        & 0.2194
        & \textbf{0.0484}
        & 0.8081 \\

        \bottomrule
    \end{tabular}
\end{table*}

\subsection{Transformer Layer Ablation}
\label{sec:layer_ablation}

We ablate how the learned visual-dial directions are distributed across
transformer blocks on Wan~2.1-1.3B, which contains 30 DiT layers.
We compare five configurations:
(1) a separate learned direction for every layer,
(2) one direction shared across all layers,
and directions applied only to
(3) early,
(4) middle, or
(5) late transformer layers.

We evaluate three concept--attribute pairs, with 10 prompts and five slider
strengths for each pair, resulting in 150 generated videos per configuration.
Table~\ref{tab:layer_ablation} reports the slider-based evaluation metrics.

The full layer-specific configuration achieves the best overall score,
while using only the middle layers performs comparably and substantially
better than using only early or late layers.
This suggests that middle transformer blocks carry much of the effective
attribute-control signal, while allowing each layer to learn its own
direction across the full network provides the strongest overall control.

A single direction shared across all layers is notably weaker, supporting
our use of layer-specific visual-dial directions.


\subsection{Motion-Control Ablations}
\label{sec:motion_ablation}

We further ablate three design choices for learning motion directions:
the source of motion-training data, the speed scaling factor $\gamma$,
and the construction of the motion target.
Since these experiments specifically evaluate motion controllability, we
report VLM-based metrics that directly assess monotonicity, editing range,
edit quality, and identity preservation.
For each experiment, we additionally report the average of these four
metrics as the overall score.


\noindent\textbf{Motion-training data.}
We first study whether the choice of motion-training data affects the
learned direction.
We compare our default green-screen training videos with general videos
containing natural backgrounds.

\begin{table}[t]
    \centering
    \small
    \caption{
        \textbf{Ablation on motion-training data.}
        We compare green-screen training videos with general videos.
    }
    \label{tab:greenscreen_ablation}

    \resizebox{\linewidth}{!}{
    \begin{tabular}{lccccc}
        \toprule
        \textbf{Training Data}
        & \textbf{Mono.} $\uparrow$
        & \textbf{Range} $\uparrow$
        & \textbf{Quality} $\uparrow$
        & \textbf{ID} $\uparrow$
        & \textbf{Overall} $\uparrow$ \\
        \midrule

        Green-screen
        & \textbf{3.93}
        & \textbf{3.47}
        & \textbf{4.70}
        & \textbf{4.47}
        & \textbf{4.14} \\

        General video
        & 3.80
        & 3.30
        & 4.57
        & 4.33
        & 4.00 \\

        \bottomrule
    \end{tabular}
    }
\end{table}

Training on green-screen videos consistently improves all four metrics,
increasing the overall score from 4.00 to 4.14.
Removing complex background variation provides cleaner supervision for
foreground motion and allows the learned direction to focus more directly
on motion magnitude.


\noindent\textbf{Speed scaling factor.}
We next vary the speed scaling factor $\gamma$ while keeping all other
training settings unchanged.
Our default setting uses $\gamma=0.5$.

\begin{table}[t]
    \centering
    \small
    \caption{
        \textbf{Ablation on the speed scaling factor $\gamma$.}
        We use $\gamma=0.5$ as the default setting.
    }
    \label{tab:speed_scale_ablation}

    \resizebox{\linewidth}{!}{
    \begin{tabular}{lccccc}
        \toprule
        $\boldsymbol{\gamma}$
        & \textbf{Mono.} $\uparrow$
        & \textbf{Range} $\uparrow$
        & \textbf{Quality} $\uparrow$
        & \textbf{ID} $\uparrow$
        & \textbf{Overall} $\uparrow$ \\
        \midrule

        0.3
        & 2.83
        & 3.03
        & \textbf{4.70}
        & 4.53
        & 3.77 \\

        0.5
        & \textbf{3.37}
        & \textbf{3.37}
        & 4.63
        & 4.43
        & \textbf{3.95} \\

        0.8
        & 2.63
        & 2.80
        & \textbf{4.70}
        & \textbf{4.57}
        & 3.68 \\

        \bottomrule
    \end{tabular}
    }
\end{table}

The intermediate value $\gamma=0.5$ achieves the highest monotonicity,
editing range, and overall score.
Although smaller or larger values slightly improve individual quality or
identity-preservation scores, they produce weaker slider progression.
We therefore use $\gamma=0.5$ as our default setting.


\noindent\textbf{Motion target.}
Finally, we compare our self-scaled motion target with a variant that
directly scales the ground-truth motion target.

\begin{table}[t]
    \centering
    \small
    \caption{
        \textbf{Ablation on the motion target.}
        We compare the proposed self-scaled target with direct scaling
        of the ground-truth motion target.
    }
    \label{tab:motion_target_ablation}

    \resizebox{\linewidth}{!}{
    \begin{tabular}{lccccc}
        \toprule
        \textbf{Motion Target}
        & \textbf{Mono.} $\uparrow$
        & \textbf{Range} $\uparrow$
        & \textbf{Quality} $\uparrow$
        & \textbf{ID} $\uparrow$
        & \textbf{Overall} $\uparrow$ \\
        \midrule

        Self-scaled (ours)
        & \textbf{3.87}
        & \textbf{3.40}
        & 4.63
        & 4.50
        & \textbf{4.10} \\

        Ground-truth-scaled
        & 1.73
        & 2.20
        & \textbf{4.80}
        & \textbf{4.70}
        & 3.36 \\

        \bottomrule
    \end{tabular}
    }
\end{table}

The self-scaled target substantially improves motion controllability,
increasing the overall score from 3.36 to 4.10.
The ground-truth-scaled variant achieves slightly higher edit quality and
identity preservation, but exhibits much weaker monotonicity and editing
range.

This suggests that directly scaling the ground-truth target tends to preserve
the original video while failing to produce a sufficiently progressive
motion slider, whereas the proposed self-scaled objective provides a stronger
control signal.



\section{Additional Method Details}
\label{sec:additional_method_details}


\subsection{Effect of Posterior Refinement}
\label{sec:posterior_refinement}

\begin{figure*}[t]
    \centering
    \includegraphics[width=\textwidth]
    {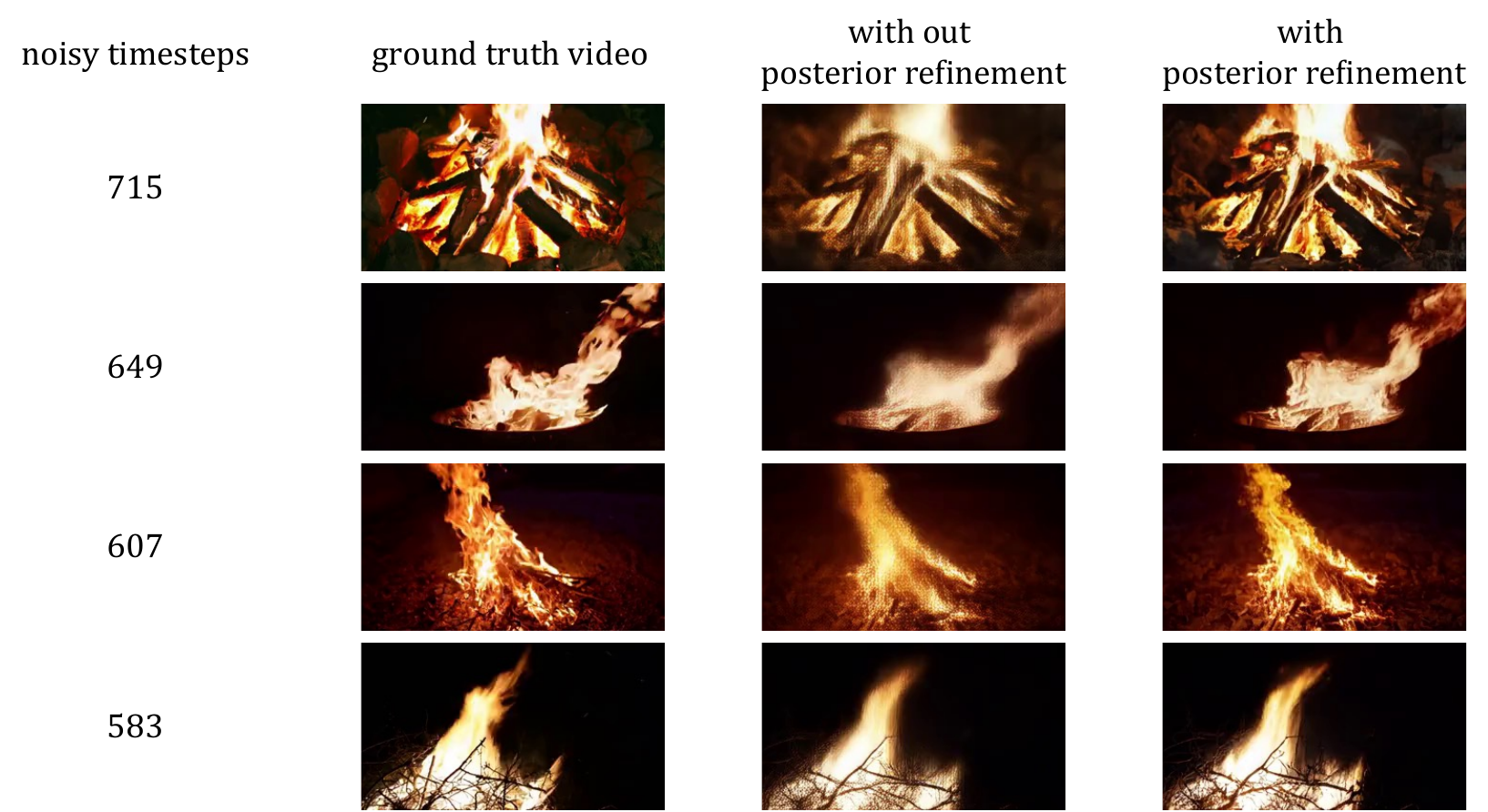}
    \caption{
        Qualitative comparisons at several noisy timesteps during training.
        The second column shows the ground-truth training video,
        the third column shows the one-step prediction of the clean video
        from the noisy latent, and the fourth column shows the corresponding
        prediction after posterior refinement.
    }
    \label{fig:posterior}
\end{figure*}

We further study the effect of posterior refinement during training.
Intuitively, during reverse diffusion, the model's one-step prediction at
high-noise timesteps can deviate from the desired posterior trajectory,
often leading to blurry structures, weakened details, or distorted content
in the reconstructed result.

Such noisy and overly smoothed predictions are also likely to be
out-of-distribution for downstream understanding models, which are typically
trained on cleaner and more natural visual inputs.
As a result, when these predicted clean videos are further processed by
understanding models such as
InternVideo2~\cite{wang2024internvideo2}
or DINOv2~\cite{oquab2024dinov2},
the extracted features and resulting supervision signals can become less
reliable and more noisy.

Fig.~\ref{fig:posterior} shows qualitative comparisons at several noisy
timesteps during training.
The second column shows the ground-truth video frames used during training.
The third column shows the one-step prediction of the clean video from noisy
latents at different timesteps.
The fourth column shows the corresponding predicted clean video after
applying posterior refinement.

Without posterior refinement, the decoded intermediate samples tend to be
over-smoothed and less structurally faithful, especially for fine-grained
appearance details such as flame boundaries, high-frequency textures, and
local shape variations.

In contrast, with posterior refinement, the intermediate predictions remain
noticeably closer to the ground-truth video, yielding sharper flame
structures, better local contrast, and more stable spatial layouts.
This leads to more meaningful feedback from the understanding models and
helps stabilize the gradients during training.


\subsection{Lucas--Kanade Optical Flow on DINOv2 Patch Features}
\label{sec:dinov2_lk_flow}

For dynamic attributes, the main paper defines the motion objective through
a flow extractor $\mathbf{m}(\cdot)$ operating in an understanding space.
In our implementation,
$\mathbf{m}(\cdot)$ is instantiated as Lucas--Kanade optical flow computed
on DINOv2 patch features rather than on RGB pixels.
This design follows the motion-magnitude scaling formulation in the main
paper.

Given a predicted video $\hat{x}_0^{ref}(\Delta)$,
we first sample a fixed set of keyframes and resize each frame to
$224 \times 224$ before feeding them into a frozen DINOv2 encoder
$\mathcal{D}$.

For each frame, we extract the normalized patch tokens,

\begin{equation*}
\mathcal{D}\!\left(\hat{x}_0^{ref}(\Delta)\right)
\in\mathbb{R}^{K\times N\times D},
\end{equation*}
where $K$ is the number of sampled frames,
$N$ is the number of spatial patches,
and $D$ is the patch-feature dimension.
We then reshape the patch tokens into a spatial grid of size
$H_p\times W_p$ with $H_pW_p=N$.

We define
$\mathbf{m}\!\left(\hat{x}_0^{ref}(\Delta)\right)$
as the patch-level flow field estimated from these DINOv2 features using
a multi-channel Lucas--Kanade formulation.

For two consecutive frames, let
$\mathbf{f}_t(i,j)\in\mathbb{R}^{D}$
denote the DINOv2 feature vector at patch location $(i,j)$.
We compute spatial gradients by central differences and temporal gradients
by frame differences:

\begin{equation*}
\mathbf{I}_x =
\frac{
\mathbf{f}_t(i,j+1)-\mathbf{f}_t(i,j-1)
}{2},
\end{equation*}
\begin{equation*}
\mathbf{I}_y =
\frac{
\mathbf{f}_t(i+1,j)-\mathbf{f}_t(i-1,j)
}{2},
\end{equation*}
\begin{equation*}
\mathbf{I}_t =
\mathbf{f}_{t+1}(i,j)-\mathbf{f}_t(i,j).
\end{equation*}
Under the Lucas--Kanade assumption, the local flow $(u,v)$ is obtained by
solving a least-squares system jointly over all feature channels.
In practice, we first form the structure-tensor terms

\begin{equation*}
I_{xx}=\sum_d I_x^{(d)}I_x^{(d)},
\end{equation*}
\begin{equation*}
I_{yy}=\sum_d I_y^{(d)}I_y^{(d)},
\end{equation*}
\begin{equation*}
I_{xy}=\sum_d I_x^{(d)}I_y^{(d)},
\end{equation*}
\begin{equation*}
I_{xt}=\sum_d I_x^{(d)}I_t^{(d)},
\qquad
I_{yt}=\sum_d I_y^{(d)}I_t^{(d)},
\end{equation*}
average them within a local spatial window, and then solve the resulting
$2\times2$ linear system independently at each patch location.

This produces a patch-level flow field

\begin{equation*}
\mathbf{m}\!\left(\hat{x}_0^{ref}(\Delta)\right)
\in\mathbb{R}^{(K-1)\times N\times 2}.
\end{equation*}
Our implementation uses replicate padding for boundary handling,
central differences for spatial gradients,
a small square averaging window for local coherence,
and Cramer's rule to solve the $2\times2$ system.

To enforce motion magnitude scaling, we follow the stop-gradient design
in the main paper and construct the target flow by scaling a detached
reference flow:

\begin{equation*}
\mathbf{m}_{\mathrm{ref}}
=
\gamma\cdot
\Bigl[
\mathbf{m}\!\left(\hat{x}_0^{ref}(\Delta)\right)
.\mathrm{sg}()
\Bigr],
\end{equation*}
where $\gamma$ is the motion scaling factor.
The dynamic loss is then

\begin{equation*}
\mathcal{L}_{\mathrm{dyn}}
=
\left\|
\mathbf{m}\!\left(\hat{x}_0^{ref}(\Delta)\right)
-
\gamma\cdot
\Bigl[
\mathbf{m}\!\left(\hat{x}_0^{ref}(\Delta)\right)
.\mathrm{sg}()
\Bigr]
\right\|_2^2.
\end{equation*}
In implementation, the detached branch is computed without gradient,
while the flow extracted from the current prediction remains differentiable
with respect to the generated video frames.


\subsection{Implementation Details on Wan 2.1}
\label{sec:wan}

We use Wan 2.1 T2V 1.3B~\cite{wan2025wanopenadvancedlargescale}
as the base video generation model.

Consistent with the formulation in the main paper,
TokenDial is implemented by injecting additive token offsets into
intermediate video patch tokens of the pretrained video DiT.

On Wan 2.1, we realize this intervention inside each DiT block by adding
the learned offset to the self-attention residual branch.
Concretely, after the hidden tokens are normalized and modulated by the
timestep-conditioned adaptive parameters, they are passed through the
self-attention layer;
the predicted token offset is then added to the self-attention output before
the gated residual update.
The subsequent cross-attention and feed-forward branches are kept unchanged.

Our choice of the self-attention residual branch is motivated by both
architectural considerations and empirical ablations.
We experimented with injecting the token offset at several alternative
locations, including before and after the modulation layers,
in the cross-attention branch, and after each full DiT block.

Among these choices, injecting into the self-attention residual branch
consistently provided the best trade-off between editability and preservation.
In particular, this location better preserves the original video structure
while still enabling meaningful attribute-level edits.

Intuitively, the self-attention branch primarily operates on the model's
latent visual tokens and thus offers a natural intervention point for
modifying internal visual semantics without excessively disturbing the
text-conditioning pathway or the overall block computation.
By contrast, injecting into cross-attention or after the full block tends
to produce less stable edits or weaker structural preservation.

This implementation preserves the core design of our method,
namely, learning attribute-specific additive offsets in the intermediate
token space rather than modifying backbone weights,
while adapting it to the architectural structure of Wan 2.1.

\FloatBarrier


\section{Additional Qualitative Results}
\label{sec:more_results}

More results for appearance sliders and motion sliders can be found in the
attached video presentation and the \texttt{index.html} file inside the
project-page folder.

\begin{figure*}[t]
    \centering
    \includegraphics[width=\textwidth]
    {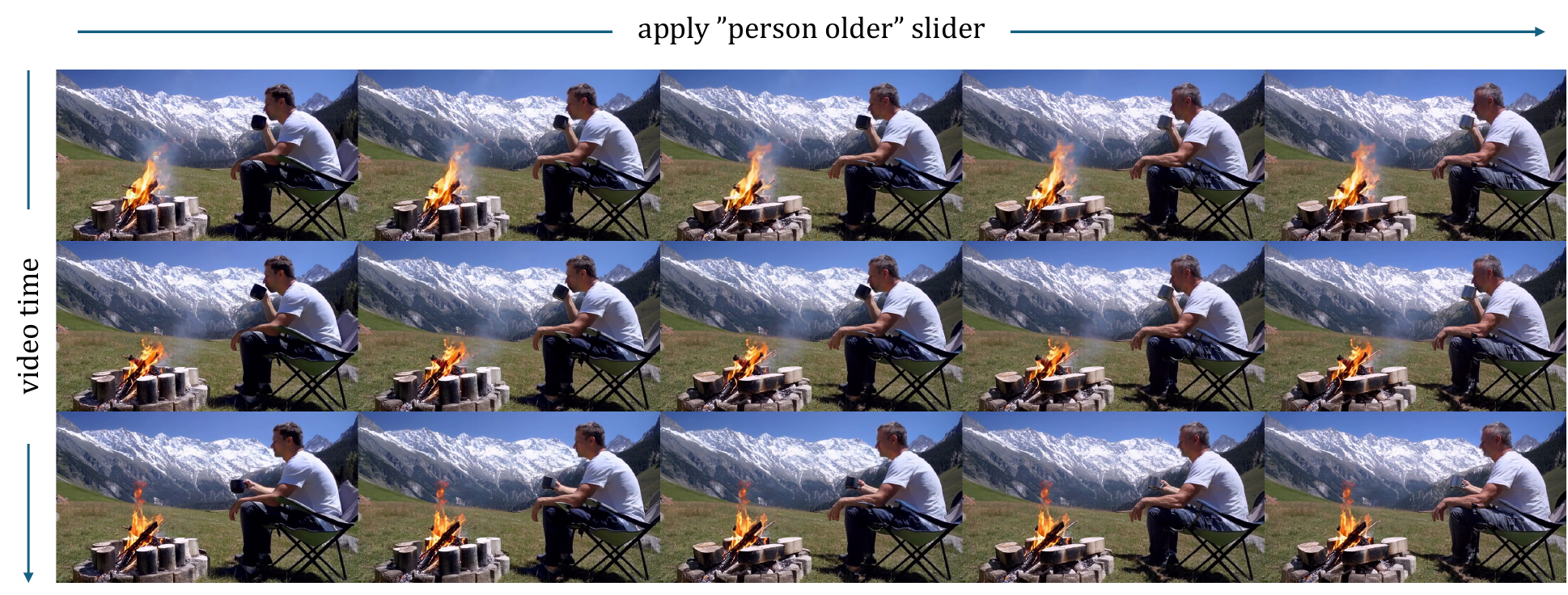}
    \caption{
        Additional results for appearance sliders.
    }
    \label{fig:more_results1}
\end{figure*}

\begin{figure*}[t]
    \centering
    \includegraphics[width=\textwidth]
    {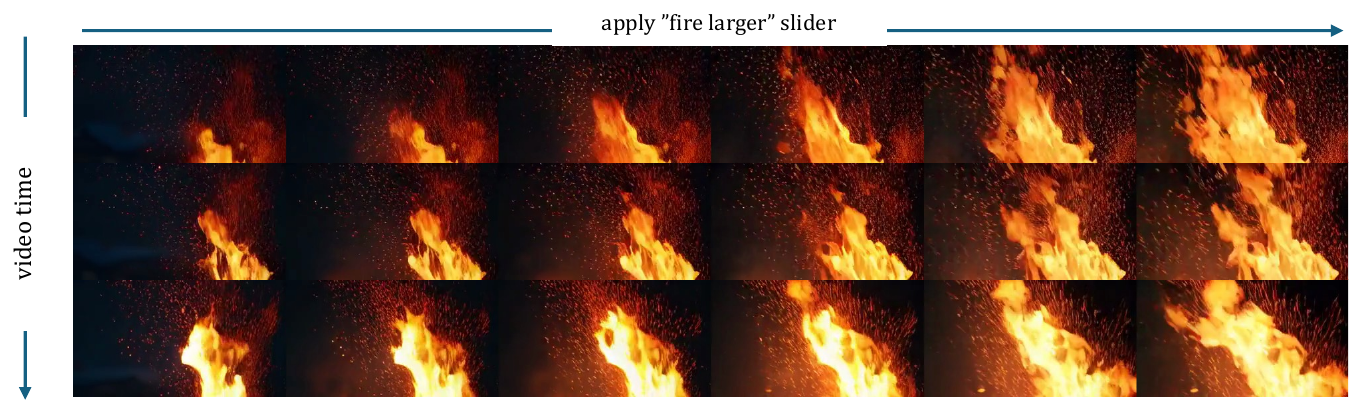}
    \caption{
        Additional results for appearance sliders.
    }
    \label{fig:more_results3}
\end{figure*}

\begin{figure*}[t]
    \centering
    \includegraphics[width=\textwidth]
    {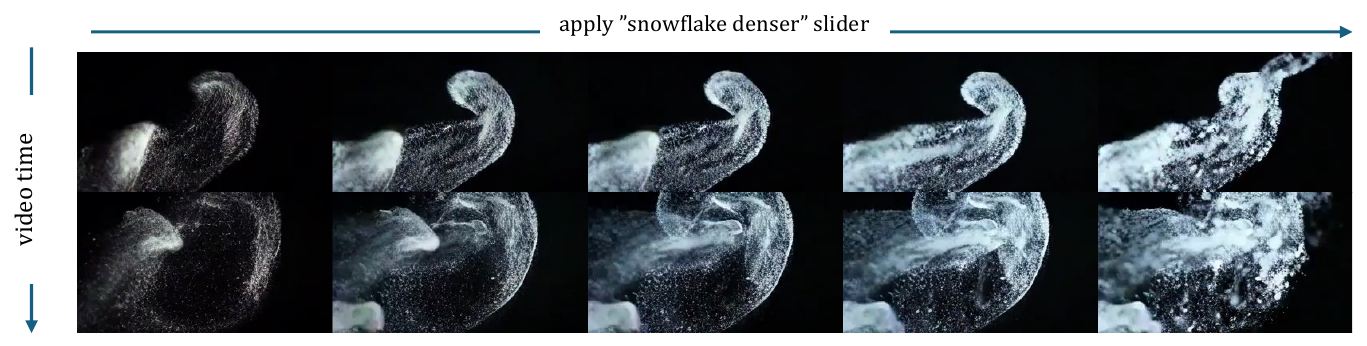}
    \caption{
        Additional results for appearance sliders.
    }
    \label{fig:more_results4}
\end{figure*}

\FloatBarrier

%% file: supplimentary/tables/human_survey.tex
\begin{table}[t]
\footnotesize
  \centering
  \caption{User study results across 6 baselines including appearance/motion sliders.}
  \label{tab:suppl_human_eval}
  \setlength{\tabcolsep}{5pt}
  \renewcommand{\arraystretch}{0.95}
  \resizebox{\linewidth}{!}{
  \begin{tabular}{lcccccc}
    \toprule
    \cmidrule(lr){2-7}
    \textbf{Method}
    & \textbf{\makecell{Editing\\Qual.}} $\uparrow$
    & \textbf{\makecell{Content\\Prev.}} $\uparrow$
    & \textbf{\makecell{BG.\\Prev.}} $\uparrow$
    & \textbf{Continuity} $\uparrow$
    & \textbf{\makecell{Less\\Artifact}} $\uparrow$
    & \textbf{\makecell{General\\Preference}} $\uparrow$ \\
    \midrule
    Concept Slider & 2.98 & 3.00 & 2.46 & 3.25 & 2.97 & 3.16 \\
    SliderSpace   & 2.89 & 2.90 & 2.51 & 3.24 & 3.12 & 3.19 \\
    FreeSliders    & 3.25 & 3.48 & 3.21 & 3.28 & 3.52 & 3.32 \\
    Text Slider    & 2.09 & \textbf{4.23} & \textbf{4.05} & 2.36 & \textbf{4.02} & 2.89 \\
    \midrule
    Senorita      & 2.73 & 4.22 & 3.95 & 2.71 & 3.71 & 2.84 \\
    UniVideo      & 3.06 & 3.99 & 3.63 & 2.92 & 3.43 & 3.09 \\
    \midrule
    Ours          & \textbf{3.91} & 4.06 & 3.59 & \textbf{3.75} & 3.68 & \textbf{3.69} \\
    \bottomrule
  \end{tabular}
  }
\end{table}

%% file: main_bibliography.bib
@String{Computer = "{IEEE} Computer" }

@String{Springer = "Springer-Verlag" }

@inproceedings{radford2021clip,
  title={Learning transferable visual models from natural language supervision},
  author={Radford, Alec and Kim, Jong Wook and Hallacy, Chris and Ramesh, Aditya and Goh, Gabriel and Agarwal, Sandhini and Sastry, Girish and Askell, Amanda and Mishkin, Pamela and Clark, Jack and others},
  booktitle={International Conference on Machine Learning},
  pages={8748--8763},
  year={2021},
  organization={PMLR}
}

@inproceedings{
kwon2023diffusion,
title={Diffusion Models Already Have A Semantic Latent Space},
author={Mingi Kwon and Jaeseok Jeong and Youngjung Uh},
booktitle={The Eleventh International Conference on Learning Representations },
year={2023},
url={https://openreview.net/forum?id=pd1P2eUBVfq}
}

@misc{park2023unsuperviseddiscoverysemanticlatent,
      title={Unsupervised Discovery of Semantic Latent Directions in Diffusion Models}, 
      author={Yong-Hyun Park and Mingi Kwon and Junghyo Jo and Youngjung Uh},
      year={2023},
      eprint={2302.12469},
      archivePrefix={arXiv},
      primaryClass={cs.CV},
      url={https://arxiv.org/abs/2302.12469}, 
}

@inproceedings{dalva2024noiseclr,
  title={NoiseCLR: A contrastive learning approach for unsupervised discovery of interpretable directions in diffusion models},
  author={Dalva, Yusuf and Yanardag, Pinar},
  booktitle={2024 IEEE/CVF Conference on Computer Vision and Pattern Recognition (CVPR)},
  pages={24209--24218},
  year={2024},
  organization={IEEE}
}

@inproceedings{
yu2025repa,
title={Representation Alignment for Generation: Training Diffusion Transformers Is Easier Than You Think},
author={Sihyun Yu and Sangkyung Kwak and Huiwon Jang and Jongheon Jeong and Jonathan Huang and Jinwoo Shin and Saining Xie},
booktitle={The Thirteenth International Conference on Learning Representations},
year={2025},
url={https://openreview.net/forum?id=DJSZGGZYVi}
}

@inproceedings{liu2024scoft,
  title={SCoFT: Self-contrastive fine-tuning for equitable image generation},
  author={Liu, Zhixuan and Schaldenbrand, Peter and Okogwu, Beverley-Claire and Peng, Wenxuan and Yun, Youngsik and Hundt, Andrew and Kim, Jihie and Oh, Jean},
  booktitle={Proceedings of the IEEE/CVF Conference on Computer Vision and Pattern Recognition},
  pages={10822--10832},
  year={2024}
}

@inproceedings{wang2024internvideo2,
  title={InternVideo2: Scaling foundation models for multimodal video understanding},
  author={Wang, Yi and Li, Kunchang and Li, Xinhao and Yu, Jiashuo and He, Yinan and Chen, Guo and Pei, Baoqi and Zheng, Rongkun and Wang, Zun and Shi, Yansong and others},
  booktitle={European Conference on Computer Vision},
  pages={396--416},
  year={2024},
  organization={Springer}
}

@article{
oquab2024dinov2,
title={{DINO}v2: Learning Robust Visual Features without Supervision},
author={Maxime Oquab and Timoth{\'e}e Darcet and Th{\'e}o Moutakanni and Huy V. Vo and Marc Szafraniec and Vasil Khalidov and Pierre Fernandez and Daniel Haziza and Francisco Massa and Alaaeldin El-Nouby and Mido Assran and Nicolas Ballas and Wojciech Galuba and Russell Howes and Po-Yao Huang and Shang-Wen Li and Ishan Misra and Michael Rabbat and Vasu Sharma and Gabriel Synnaeve and Hu Xu and Herve Jegou and Julien Mairal and Patrick Labatut and Armand Joulin and Piotr Bojanowski},
journal={Transactions on Machine Learning Research},
issn={2835-8856},
year={2024},
url={https://openreview.net/forum?id=a68SUt6zFt},
}

@inproceedings{yang2025cogvideox,
  title={CogVideoX: Text-to-video diffusion models with an expert transformer},
  author={Yang, Zhuoyi and Teng, Jiayan and Zheng, Wendi and Ding, Ming and Huang, Shiyu and Xu, Jiazheng and Yang, Yuanming and Hong, Wenyi and Zhang, Xiaohan and Feng, Guanyu and others},
  booktitle={International Conference on Learning Representations},
  volume={2025},
  pages={83048--83077},
  year={2025}
}

@misc{wan2025wanopenadvancedlargescale,
      title={Wan: Open and Advanced Large-Scale Video Generative Models}, 
      author={Team Wan and Ang Wang and Baole Ai and Bin Wen and Chaojie Mao and Chen-Wei Xie and Di Chen and Feiwu Yu and Haiming Zhao and Jianxiao Yang and Jianyuan Zeng and Jiayu Wang and Jingfeng Zhang and Jingren Zhou and Jinkai Wang and Jixuan Chen and Kai Zhu and Kang Zhao and Keyu Yan and Lianghua Huang and Mengyang Feng and Ningyi Zhang and Pandeng Li and Pingyu Wu and Ruihang Chu and Ruili Feng and Shiwei Zhang and Siyang Sun and Tao Fang and Tianxing Wang and Tianyi Gui and Tingyu Weng and Tong Shen and Wei Lin and Wei Wang and Wei Wang and Wenmeng Zhou and Wente Wang and Wenting Shen and Wenyuan Yu and Xianzhong Shi and Xiaoming Huang and Xin Xu and Yan Kou and Yangyu Lv and Yifei Li and Yijing Liu and Yiming Wang and Yingya Zhang and Yitong Huang and Yong Li and You Wu and Yu Liu and Yulin Pan and Yun Zheng and Yuntao Hong and Yupeng Shi and Yutong Feng and Zeyinzi Jiang and Zhen Han and Zhi-Fan Wu and Ziyu Liu},
      year={2025},
      eprint={2503.20314},
      archivePrefix={arXiv},
      primaryClass={cs.CV},
      url={https://arxiv.org/abs/2503.20314}, 
}

@article{ezra2025freesliders,
  title={FreeSliders: Training-Free, Modality-Agnostic Concept Sliders for Fine-Grained Diffusion Control in Images, Audio, and Video},
  author={Ezra, Rotem and Zisling, Hedi and Berman, Nimrod and Naiman, Ilan and Gorkor, Alexey and Nochumsohn, Liran and Nachmani, Eliya and Azencot, Omri},
  journal={arXiv preprint arXiv:2511.00103},
  year={2025}
}

@inproceedings{chiu2026text,
  title={Text Slider: Efficient and Plug-and-Play Continuous Concept Control for Image/Video Synthesis via LoRA Adapters},
  author={Chiu, Pin-Yen and Fang, I and Chen, Jun-Cheng and others},
  booktitle={Proceedings of the IEEE/CVF Winter Conference on Applications of Computer Vision},
  pages={613--622},
  year={2026}
}

@inproceedings{gandikota2024concept,
  title={Concept Sliders: LoRA adaptors for precise control in diffusion models},
  author={Gandikota, Rohit and Materzy{\'n}ska, Joanna and Zhou, Tingrui and Torralba, Antonio and Bau, David},
  booktitle={European Conference on Computer Vision},
  pages={172--188},
  year={2024},
  organization={Springer}
}

@inproceedings{gandikota2025sliderspace,
  title={SliderSpace: Decomposing the visual capabilities of diffusion models},
  author={Gandikota, Rohit and Wu, Zongze and Zhang, Richard and Bau, David and Shechtman, Eli and Kolkin, Nick},
  booktitle={2025 IEEE/CVF International Conference on Computer Vision (ICCV)},
  pages={15994--16003},
  year={2025},
  organization={IEEE}
}

@article{kamenetsky2025saedit,
  title={SAEdit: Token-level control for continuous image editing via Sparse AutoEncoder},
  author={Kamenetsky, Ronen and Dorfman, Sara and Garibi, Daniel and Paiss, Roni and Patashnik, Or and Cohen-Or, Daniel},
  journal={arXiv preprint arXiv:2510.05081},
  year={2025}
}

@inproceedings{parihar2025kontinuouskontextcontinuousstrength,
  title={Kontinuous Kontext: Continuous strength control for instruction-based image editing},
  author={Parihar, Rishubh and Patashnik, Or and Ostashev, Daniil and Radhakrishnan, Venkatesh Babu and Cohen-Or, Daniel and Wang, Kuan-Chieh Jackson},
  booktitle={Proceedings of the IEEE/CVF Conference on Computer Vision and Pattern Recognition},
  pages={37929--37939},
  year={2026}
}

@inproceedings{zhang2018lpips,
  title={The unreasonable effectiveness of deep features as a perceptual metric},
  author={Zhang, Richard and Isola, Phillip and Efros, Alexei A and Shechtman, Eli and Wang, Oliver},
  booktitle={2018 IEEE/CVF Conference on Computer Vision and Pattern Recognition},
  pages={586--595},
  year={2018},
  organization={IEEE}
}

@inproceedings{lucas1981,
  author    = {Lucas, Bruce D. and Kanade, Takeo},
  title     = {An iterative image registration technique with an application to stereo vision},
  booktitle = {Proceedings of the 7th International Joint Conference on Artificial Intelligence},
  year      = {1981},
  pages     = {674--679}
}

@article{garibi2025tokenverse,
  title={TokenVerse: Versatile multi-concept personalization in token modulation space},
  author={Garibi, Daniel and Yadin, Shahar and Paiss, Roni and Tov, Omer and Zada, Shiran and Ephrat, Ariel and Michaeli, Tomer and Mosseri, Inbar and Dekel, Tali},
  journal={ACM Transactions on Graphics (TOG)},
  volume={44},
  number={4},
  pages={1--11},
  year={2025},
  publisher={ACM New York, NY, USA}
}

@inproceedings{ju2026editverse,
  title={EditVerse: Unifying image and video editing and generation with in-context learning},
  author={Ju, Xuan and Wang, Tianyu and Zhou, Yuqian and Zhang, He and Liu, Qing and Zhao, Cherry and Zhang, Zhifei and Li, Yijun and Cai, Yuanhao and Liu, Shaoteng and others},
  booktitle={International Conference on Learning Representations},
  volume={2026},
  pages={137234--137255},
  year={2026}
}

@inproceedings{zhang2023controlnet,
  title={Adding conditional control to text-to-image diffusion models},
  author={Zhang, Lvmin and Rao, Anyi and Agrawala, Maneesh},
  booktitle={2023 IEEE/CVF International Conference on Computer Vision (ICCV)},
  pages={3813--3824},
  year={2023},
  organization={IEEE}
}

@article{chen2024pixart,
  title={{PixArt}-$\delta$: Fast and Controllable Image Generation with Latent Consistency Models},
  author={Chen, Junsong and Wu, Yue and Luo, Simian and Xie, Enze and Paul, Sayak and Luo, Ping and Zhao, Hang and Li, Zhenguo},
  journal={arXiv preprint arXiv:2401.05252},
  year={2024}
}

@inproceedings{ruiz2023dreambooth,
  title={DreamBooth: Fine tuning text-to-image diffusion models for subject-driven generation},
  author={Ruiz, Nataniel and Li, Yuanzhen and Jampani, Varun and Pritch, Yael and Rubinstein, Michael and Aberman, Kfir},
  booktitle={2023 IEEE/CVF Conference on Computer Vision and Pattern Recognition (CVPR)},
  pages={22500--22510},
  year={2023},
  organization={IEEE}
}

@misc{hu2025hunyuancustom,
      title={HunyuanCustom: A Multimodal-Driven Architecture for Customized Video Generation}, 
      author={Teng Hu and Zhentao Yu and Zhengguang Zhou and Sen Liang and Yuan Zhou and Qin Lin and Qinglin Lu},
      year={2025},
      eprint={2505.04512},
      archivePrefix={arXiv},
      primaryClass={cs.CV},
      url={https://arxiv.org/abs/2505.04512}, 
}

@inproceedings{xing2025motioncanvas,
  title={MotionCanvas: Cinematic shot design with controllable image-to-video generation},
  author={Xing, Jinbo and Mai, Long and Ham, Cusuh and Huang, Jiahui and Mahapatra, Aniruddha and Fu, Chi-Wing and Wong, Tien-Tsin and Liu, Feng},
  booktitle={Proceedings of the Special Interest Group on Computer Graphics and Interactive Techniques Conference Conference Papers},
  pages={1--11},
  year={2025}
}

@inproceedings{lee2026generative,
  title={Generative video motion editing with 3D point tracks},
  author={Lee, Yao-Chih and Zhang, Zhoutong and Huang, Jiahui and Wang, Jui-Hsien and Lee, Joon-Young and Huang, Jia-Bin and Shechtman, Eli and Li, Zhengqi},
  booktitle={Proceedings of the IEEE/CVF Conference on Computer Vision and Pattern Recognition},
  pages={18306--18318},
  year={2026}
}

@inproceedings{geng2025motion,
  title={Motion prompting: Controlling video generation with motion trajectories},
  author={Geng, Daniel and Herrmann, Charles and Hur, Junhwa and Cole, Forrester and Zhang, Serena and Pfaff, Tobias and Lopez-Guevara, Tatiana and Aytar, Yusuf and Rubinstein, Michael and Sun, Chen and others},
  booktitle={2025 IEEE/CVF Conference on Computer Vision and Pattern Recognition (CVPR)},
  pages={1--12},
  year={2025},
  organization={IEEE}
}

@inproceedings{bai2025recammaster,
  title={ReCamMaster: Camera-controlled generative rendering from a single video},
  author={Bai, Jianhong and Xia, Menghan and Fu, Xiao and Wang, Xintao and Mu, Lianrui and Cao, Jinwen and Liu, Zuozhu and Hu, Haoji and Bai, Xiang and Wan, Pengfei and others},
  booktitle={2025 IEEE/CVF International Conference on Computer Vision (ICCV)},
  pages={14834--14844},
  year={2025},
  organization={IEEE}
}

@article{mou2024revideo,
  title={ReVideo: Remake a video with motion and content control},
  author={Mou, Chong and Cao, Mingdeng and Wang, Xintao and Zhang, Zhaoyang and Shan, Ying and Zhang, Jian},
  journal={Advances in Neural Information Processing Systems},
  volume={37},
  pages={18481--18505},
  year={2024}
}

@inproceedings{zi2025senorita, title={Señorita-2M: A High-Quality Instruction-based Dataset for General Video Editing by Video Specialists}, 
      author={Bojia Zi and Penghui Ruan and Marco Chen and Xianbiao Qi and Shaozhe Hao and Shihao Zhao and Youze Huang and Bin Liang and Rong Xiao and Kam-Fai Wong}, 
      booktitle = {Advances in Neural Information Processing Systems},
      note = {Datasets and Benchmarks Track},
      year={2025}, 
}

@article{
ku2024anyv2v,
title={AnyV2V: A Tuning-Free Framework For Any Video-to-Video Editing Tasks},
author={Max Ku and Cong Wei and Weiming Ren and Huan Yang and Wenhu Chen},
journal={Transactions on Machine Learning Research},
issn={2835-8856},
year={2024},
url={https://openreview.net/forum?id=RFrJCkw2oa},
}

@misc{decart2025lucyedit,
  title   = {Lucy Edit: Open-Weight Text-Guided Video Editing},
  author  = {DecartAI Team},
  year    = {2025}
 }

@inproceedings{liu2025generative,
  title={Generative video propagation},
  author={Liu, Shaoteng and Wang, Tianyu and Wang, Jui-Hsien and Liu, Qing and Zhang, Zhifei and Lee, Joon-Young and Li, Yijun and Yu, Bei and Lin, Zhe and Kim, Soo Ye and others},
  booktitle={2025 IEEE/CVF Conference on Computer Vision and Pattern Recognition (CVPR)},
  pages={17712--17722},
  year={2025},
  organization={IEEE}
}

@article{kirstain2023pick,
  title={Pick-a-Pic: An open dataset of user preferences for text-to-image generation},
  author={Kirstain, Yuval and Polyak, Adam and Singer, Uriel and Matiana, Shahbuland and Penna, Joe and Levy, Omer},
  journal={Advances in Neural Information Processing Systems},
  volume={36},
  pages={36652--36663},
  year={2023}
}

@inproceedings{wang2023internvid,
  title={InternVid: A Large-scale Video-Text Dataset for Multimodal Understanding and Generation},
  author={Wang, Yi and He, Yinan and Li, Yizhuo and Li, Kunchang and Yu, Jiashuo and Ma, Xin and Li, Xinhao and Chen, Guo and Chen, Xinyuan and Wang, Yaohui and others},
  booktitle={The Twelfth International Conference on Learning Representations},
  year={2023}
}

@inproceedings{radford2015unsupervised,
  title={Unsupervised Representation Learning with Deep Convolutional Generative Adversarial Networks},
  author={Radford, Alec and Metz, Luke and Chintala, Soumith},
  booktitle={International Conference on Learning Representations},
  year={2016}
}

@article{chen2016infogan,
  title={InfoGAN: Interpretable representation learning by information maximizing generative adversarial nets},
  author={Chen, Xi and Duan, Yan and Houthooft, Rein and Schulman, John and Sutskever, Ilya and Abbeel, Pieter},
  journal={Advances in Neural Information Processing Systems},
  volume={29},
  year={2016}
}

@article{harkonen2020ganspace,
  title={GANSpace: Discovering interpretable GAN controls},
  author={H{\"a}rk{\"o}nen, Erik and Hertzmann, Aaron and Lehtinen, Jaakko and Paris, Sylvain},
  journal={Advances in Neural Information Processing Systems},
  volume={33},
  pages={9841--9850},
  year={2020}
}

@article{zhang2026videorepa,
  title={VideoREPA: Learning physics for video generation through relational alignment with foundation models},
  author={Zhang, Xiangdong and Liao, Jiaqi and Zhang, Shaofeng and Meng, Fanqing and Wan, Xiangpeng and Yan, Junchi and Cheng, Yu},
  journal={Advances in Neural Information Processing Systems},
  volume={38},
  pages={122647--122676},
  year={2026}
}

@inproceedings{
gal2023an,
title={An Image is Worth One Word: Personalizing Text-to-Image Generation using Textual Inversion},
author={Rinon Gal and Yuval Alaluf and Yuval Atzmon and Or Patashnik and Amit Haim Bermano and Gal Chechik and Daniel Cohen-Or},
booktitle={The Eleventh International Conference on Learning Representations },
year={2023},
url={https://openreview.net/forum?id=NAQvF08TcyG}
}

@article{ho2020denoising,
  title={Denoising diffusion probabilistic models},
  author={Ho, Jonathan and Jain, Ajay and Abbeel, Pieter},
  journal={Advances in Neural Information Processing Systems},
  volume={33},
  pages={6840--6851},
  year={2020}
}

@INPROCEEDINGS{tanjim2024mitigatebias,
  author={Tanjim, Md Mehrab and Singh, Krishna Kumar and Kafle, Kushal and Sinha, Ritwik and Cottrell, Garrison W.},
  booktitle={2024 IEEE/CVF Winter Conference on Applications of Computer Vision (WACV)}, 
  title={Discovering and Mitigating Biases in CLIP-based Image Editing}, 
  year={2024},
  volume={},
  number={},
  pages={2972-2981},
  doi={10.1109/WACV57701.2024.00296}}

@inproceedings{
hu2022lora,
title={Lo{RA}: Low-Rank Adaptation of Large Language Models},
author={Edward J Hu and Yelong Shen and Phillip Wallis and Zeyuan Allen-Zhu and Yuanzhi Li and Shean Wang and Lu Wang and Weizhu Chen},
booktitle={International Conference on Learning Representations},
year={2022},
url={https://openreview.net/forum?id=nZeVKeeFYf9}
}

@inproceedings{wei2026univideo,
  title={UniVideo: Unified understanding, generation, and editing for videos},
  author={Wei, Cong and Liu, Quande and Ye, Zixuan and Wang, Qiulin and Wang, Xintao and Wan, Pengfei and Gai, Kun and Chen, Wenhu},
  booktitle={International Conference on Learning Representations},
  volume={2026},
  pages={113905--113933},
  year={2026}
}

@misc{zheng2024opensorademocratizingefficientvideo,
      title={Open-Sora: Democratizing Efficient Video Production for All}, 
      author={Zangwei Zheng and Xiangyu Peng and Tianji Yang and Chenhui Shen and Shenggui Li and Hongxin Liu and Yukun Zhou and Tianyi Li and Yang You},
      year={2024},
      eprint={2412.20404},
      archivePrefix={arXiv},
      primaryClass={cs.CV},
      url={https://arxiv.org/abs/2412.20404}, 
}

@inproceedings{seo2026propflylearningpropagateonthefly,
  title={PropFly: Learning to Propagate via On-the-Fly Supervision from Pre-trained Video Diffusion Models},
  author={Seo, Wonyong and Moon, Jaeho and Lee, Jaehyup and Kim, Soo Ye and Kim, Munchurl},
  booktitle={Proceedings of the IEEE/CVF Conference on Computer Vision and Pattern Recognition},
  pages={43228--43238},
  year={2026}
}

@inproceedings{peebles2023scalable,
  title={Scalable diffusion models with transformers},
  author={Peebles, William and Xie, Saining},
  booktitle={Proceedings of the IEEE/CVF International Conference on Computer Vision},
  pages={4195--4205},
  year={2023}
}

@inproceedings{wolf2026continuouscontroleditingmodels,
  title={Continuous Control of Editing Models via Adaptive-Origin Guidance},
  author={Wolf, Alon and Katzir, Chen and Aberman, Kfir and Patashnik, Or},
  booktitle={Proceedings of the Special Interest Group on Computer Graphics and Interactive Techniques Conference Conference Papers},
  pages={1--12},
  year={2026}
}

@inproceedings{kulikov2025flowedit,
  title={FlowEdit: Inversion-free text-based editing using pre-trained flow models},
  author={Kulikov, Vladimir and Kleiner, Matan and Huberman-Spiegelglas, Inbar and Michaeli, Tomer},
  booktitle={2025 IEEE/CVF International Conference on Computer Vision (ICCV)},
  pages={19721--19730},
  year={2025},
  organization={IEEE}
}

@misc{barbhuiya_redcat_vecteezy,
  author = {Towfiqu Ahamed Barbhuiya},
  title  = {Red cat looking around outdoor},
  url    = {https://www.vecteezy.com/video/29214016-red-cat-looking-around-outdoor},
  year = {2025},
  note   = {Vecteezy video}
}

@misc{chunaeva_man_vecteezy,
  author       = {Kseniia Chunaeva},
  title        = {Young Caucasian man in profile, city street bokeh lights, contemplative mood during evening rush hour, urban lifestyle perspective},
  howpublished = {Vecteezy},
  url          = {https://www.vecteezy.com/video/72175010-young-caucasian-man-in-profile-city-street-bokeh-lights-contemplative-mood-during-evening-rush-hour-urban-lifestyle-perspective},
  year = {2025},
  note         = {Vecteezy video}
}

@article{comanici2025gemini,
  title={Gemini 2.5: Pushing the frontier with advanced reasoning, multimodality, long context, and next generation agentic capabilities},
  author={Comanici, Gheorghe and Bieber, Eric and Schaekermann, Mike and Pasupat, Ice and Sachdeva, Noveen and Dhillon, Inderjit and Blistein, Marcel and Ram, Ori and Zhang, Dan and Rosen, Evan and others},
  journal={arXiv preprint arXiv:2507.06261},
  year={2025}
}

@inproceedings{
bolya2026perception,
title={Perception Encoder: The best visual embeddings are not at the output of the network},
author={Daniel Bolya and Po-Yao Huang and Peize Sun and Jang Hyun Cho and Andrea Madotto and Chen Wei and Tengyu Ma and Jiale Zhi and Jathushan Rajasegaran and Hanoona Abdul Rasheed and Junke Wang and Marco Monteiro and Hu Xu and Shiyu Dong and Nikhila Ravi and Shang-Wen Li and Piotr Dollar and Christoph Feichtenhofer},
booktitle={The Thirty-ninth Annual Conference on Neural Information Processing Systems},
year={2025},
url={https://openreview.net/forum?id=INqBOmwIpG}
}
